\documentclass{article}

\usepackage{arxiv}

\usepackage[utf8]{inputenc} 
\usepackage[T1]{fontenc}    
\usepackage{hyperref}       
\usepackage{url}            
\usepackage{booktabs}       
\usepackage{amsfonts}       
\usepackage{nicefrac}       
\usepackage{microtype}      
\usepackage{lipsum}		
\usepackage{graphicx}
\usepackage{natbib}
\usepackage{doi}
\usepackage{amsmath}
\usepackage{amsthm}
\usepackage{multirow}

\usepackage{algorithm} 
\usepackage{algpseudocode}

\usepackage{subcaption} 

\theoremstyle{definition}

\title{Learning Through Energy Refinement and Manifold Projection:
A Cooperative EBM-AE Framework}

\author{ 
	{Ryad Zemouri} \\
	Hydro-Québec\\
    75, boulevard René-Lévesque Ouest\\
    Montréal, Québec, Canada\\
}

\hypersetup{
pdftitle={title},
pdfsubject={q-bio.NC, q-bio.QM},
pdfauthor={Ryad Zemouri, },
}

\begin{document}
\maketitle

\begin{abstract}
	Energy-Based Models (EBMs) provide a flexible framework for generative modeling by learning an energy landscape that assigns low energy values to realistic samples and higher energies to unlikely observations. Despite their theoretical appeal, training EBMs remains challenging due to the computational cost of Langevin sampling and the difficulty of efficiently exploring the learned data manifold. In this work, we propose a cooperative Energy-Based Model and Autoencoder (EBM-AE) framework that combines energy-based refinement with manifold projection. The proposed approach jointly trains an EBM with a denoising autoencoder and introduces an iterative EBM$\rightarrow$AE$\rightarrow$EBM sampling procedure in which Langevin dynamics and autoencoder projection alternately refine generated samples. Within this framework, the autoencoder acts as a manifold projection operator that regularizes sampling trajectories, while the EBM performs energy-based refinement toward low-energy regions of the learned distribution. Extensive experiments conducted on the MNIST dataset demonstrate that joint EBM-AE training substantially improves generation quality compared with a conventional autoencoder. 
    Beyond unconditional generation, we evaluate the proposed framework on image inpainting tasks involving structured and random masks. The results show that manifold projection provides the majority of the reconstruction capability, whereas the final energy-based refinement becomes increasingly beneficial as the reconstruction problem becomes more challenging. 
    Taken together, the results indicate that combining manifold projection and energy minimization provides an effective and interpretable framework for generation, reconstruction, and out-of-distribution detection, while offering new insights into the complementary roles of energy-based modeling and representation learning.
\end{abstract}


\section{Introduction} Learning generative models capable of producing realistic samples while accurately characterizing the underlying data distribution remains a central challenge in machine learning. Over the past decade, several families of deep generative models have emerged, including Variational Autoencoders (VAEs) \cite{Kingma2014}, Generative Adversarial Networks (GANs) \cite{Goodfellow2014}, diffusion models \cite{Ho2020} and Energy-Based Models (EBMs) \cite{LeCun2006,YilunDu2019}. While these approaches have achieved remarkable success, each of them exhibits specific limitations with respect to sample quality, diversity, training stability, or distributional robustness. Among these approaches, Energy-Based Models offer an attractive and conceptually simple framework. Rather than directly learning a generator, an EBM learns an energy function that assigns low energy values to samples belonging to the data distribution and higher energy values to unlikely observations. Sample generation is subsequently performed through an optimization process, typically based on Langevin dynamics, that progressively moves samples toward low-energy regions of the learned landscape \cite{LeCun2006,YilunDu2019}. This formulation provides several desirable properties, including implicit generation, flexibility of the learned distribution, and strong capabilities for out-of-distribution (OOD) detection. Despite these advantages, training EBMs remains challenging. Efficient sampling often requires a large number of Langevin iterations, replay-buffer mechanisms, and careful regularization strategies in order to maintain stable energy landscapes \cite{Tieleman2008,YilunDu2019}. Moreover, generated samples may remain affected by local artifacts or may explore regions that are only partially consistent with the underlying data manifold. In parallel, autoencoders have demonstrated a strong ability to capture low-dimensional representations of complex datasets and to project noisy observations back onto meaningful data manifolds \cite{Vincent2010}. Denoising autoencoders, in particular, learn reconstruction operators that remove perturbations while preserving the semantic structure of the data. However, conventional autoencoders generally lack an explicit probabilistic or energy-based mechanism capable of distinguishing valid and invalid samples. 

Motivated by the complementary strengths of these two families of models, a cooperative framework that jointly trains an Energy-Based Model and a denoising autoencoder is proposed. The central idea is to alternate manifold projection and energy-based refinement through a cooperative EBM$\rightarrow$AE$\rightarrow$EBM sampling procedure. Within this framework, the autoencoder acts as a manifold projection operator that regularizes the trajectories generated by Langevin dynamics, while the EBM refines projected samples toward low-energy and high-density regions of the learned distribution. The proposed framework produces several desirable properties. First, it substantially improves generation quality compared with a conventional autoencoder. Second, it naturally supports image reconstruction and inpainting through iterative refinement of corrupted observations. Third, the learned energy function retains strong OOD detection capabilities while remaining compatible with high-quality sample generation. The main contributions of this work are summarized as follows: 

\begin{itemize} 
\item A cooperative EBM-AE framework that alternates energy minimization and manifold projection during sampling is proposed. 
\item A joint training strategy is introduced in which the autoencoder is trained directly on replay-buffer samples generated by the EBM, allowing it to learn a projection operator adapted to the energy landscape. 
\item Extensive experiments on MNIST are presented where jointly training the EBM and the autoencoder substantially improves generation quality, sample validity, and image reconstruction performance. 
 \end{itemize}

\section{Proposed EBM-AE Framework} 
The proposed framework combines an Energy-Based Model (EBM) and a denoising autoencoder (AE) within a unified cooperative learning architecture. The EBM learns an energy landscape that assigns low energy values to realistic samples and higher energies to unlikely observations, while the autoencoder learns a projection toward the underlying data manifold. By alternating energy-based refinement and manifold projection, the framework progressively transforms noisy samples into realistic low-energy configurations. Unlike conventional EBM training approaches that rely exclusively on Langevin dynamics, the proposed framework periodically projects intermediate samples through the autoencoder. This additional projection step improves replay-buffer quality, stabilizes sampling trajectories, and facilitates exploration of the learned data manifold. 

\subsection{Energy-Based Model} 

Let $x \in \mathbb{R}^{d}$ denote an input image. The Energy-Based Model defines a scalar-valued energy function \begin{equation} E_{\theta}(x): \mathbb{R}^{d}\rightarrow \mathbb{R} \label{equ:energy} \end{equation} parameterized by $\theta$. Realistic samples are assigned low energy values, whereas unlikely or out-of-distribution observations receive higher energies. The energy function implicitly defines the probability density \begin{equation} p_{\theta}(x) = \frac{\exp\!\left(-E_{\theta}(x)\right)} {Z_{\theta}} \label{equ:probability_density} \end{equation} where \begin{equation} Z_{\theta} = \int \exp\!\left(-E_{\theta}(x)\right)\,dx \label{equ:partition_function} \end{equation} denotes the partition function. Training aims at decreasing the energy of real samples while increasing the energy of generated samples. Let $x^{+}$ denote a minibatch of positive samples drawn from the training set and $x^{-}$ a minibatch of negative samples obtained from the replay buffer. The EBM objective is defined as \begin{equation} \mathcal{L}_{EBM} = \mathbb{E}\!\left[E_{\theta}(x^{+})\right] - \mathbb{E}\!\left[E_{\theta}(x^{-})\right] + \alpha \mathcal{R}_{E} \label{equ:EBM_objective} \end{equation} where \begin{equation} \mathcal{R}_{E} = \mathbb{E}\!\left[E_{\theta}(x^{+})^{2}\right] + \mathbb{E}\!\left[E_{\theta}(x^{-})^{2}\right] \label{equ:energy_regularization} \end{equation} is an energy regularization term introduced to prevent uncontrolled growth of energy values and improve training stability. The coefficient $\alpha$ controls the strength of this regularization. 

\subsection{Joint Autoencoder} 

The autoencoder consists of an encoder \begin{equation} z=f_{\phi}(x) \end{equation} and a decoder \begin{equation} \hat{x}=g_{\phi}(z), \end{equation} where $\phi$ denotes the parameters of both networks. Unlike a conventional denoising autoencoder, the proposed AE is not trained exclusively on real samples. Instead, it is trained on samples produced by the EBM sampling process. More precisely, replay-buffer samples are perturbed with Gaussian noise and subsequently reconstructed by the encoder-decoder architecture. The autoencoder objective combines a standard reconstruction term with an energy-based regularization term: \begin{equation} \mathcal{L}_{AE} = \mathcal{L}_{rec} + \beta \mathcal{L}_{energy} \label{equ:AE_objective} \end{equation} where \begin{equation} \mathcal{L}_{rec} = \|x-\hat{x}\|_2^2 \label{equ:AE_rec_loss} \end{equation} and \begin{equation} \mathcal{L}_{energy} = E_{\theta}(\hat{x}). \label{equ:AE_energy_loss} \end{equation} The coefficient $\beta$ controls the contribution of the energy term. This objective encourages the autoencoder not only to reconstruct its inputs accurately, but also to project them toward regions of low energy. As a consequence, the autoencoder learns an implicit manifold projection operator that maps noisy or off-manifold samples toward regions that are more consistent with the learned data distribution. 

\subsection{Cooperative EBM-AE-EBM Sampling} 
A key component of the proposed framework is the cooperative interaction between Langevin dynamics and manifold projection. Starting from an initial sample $x_t$, Langevin dynamics perform an energy-minimization step according to \begin{equation} x_{t+1} = x_t - \eta \nabla_x E_{\theta}(x_t) + \sigma \epsilon_t \label{equ:langevin_dynamics} \end{equation} where $\eta$ denotes the step size, $\sigma$ controls the injected Gaussian noise, and $\epsilon_t\sim\mathcal{N}(0,I)$. This operation moves samples toward lower-energy regions of the learned landscape. After $K$ Langevin steps, an additional Gaussian perturbation is applied: \begin{equation} \tilde{x}_{t+K} = x_{t+K} + \sigma\epsilon . \end{equation} The perturbed sample is then projected through the autoencoder: \begin{equation} \hat{x}_{t+K} = g_{\phi} \Big( f_{\phi}(\tilde{x}_{t+K}) \Big) \label{equ:AE_projection} \end{equation} The projection stage encourages samples to move toward the low-dimensional data manifold learned by the jointly trained autoencoder. In practice, it removes sampling artifacts and improves the quality of replay-buffer samples before the subsequent refinement stage. Finally, an additional Langevin refinement stage composed of $K$ sampling steps is applied to $\hat{x}_{t+K}$. The complete cooperative cycle can therefore be summarized as \begin{equation} x \;\xrightarrow{\text{EBM}}\; x' \;\xrightarrow{\text{AE}}\; x'' \;\xrightarrow{\text{EBM}}\; x''' \label{equ:complete_EBM_AE} \end{equation} This alternating projection-refinement mechanism constitutes the core contribution of the proposed framework. The autoencoder regularizes Langevin trajectories by projecting samples onto the learned manifold, while the EBM subsequently refines these projections toward lower-energy regions. The EBM$\rightarrow$AE$\rightarrow$EBM cycle constitutes the fundamental building block of the proposed framework and may be repeated multiple times during generation or reconstruction. 

\subsection{Joint Training Procedure} 

The proposed framework is trained in two successive stages. During the first stage, the autoencoder is pretrained as a conventional denoising autoencoder using real training samples. This initialization provides a meaningful approximation of the data manifold and prevents replay-buffer samples from drifting toward unrealistic regions during the early stages of joint optimization. Empirically, this warm-up phase significantly accelerates convergence and improves training stability. During the second stage, the EBM and the autoencoder are trained jointly. Following \cite{YilunDu2019}, a replay buffer $\mathcal{B}$ is maintained throughout training to store previously generated samples. At the beginning of training, the replay buffer is initialized with samples drawn from a uniform distribution. At each training iteration, the initial batch is obtained by drawing approximately $95\%$ of the samples from the replay buffer and $5\%$ from a uniform noise distribution: \begin{equation} x_0 \sim 0.95\,\mathcal{B} + 0.05\,U(-1,1). \end{equation} The resulting minibatch is then refined through the cooperative EBM$\rightarrow$AE$\rightarrow$EBM procedure described in Equation~\ref{equ:complete_EBM_AE}, producing a batch of negative samples $x^{-}$. Unlike conventional EBM approaches, negative samples are therefore not generated directly from random noise but from replay-buffer samples that are iteratively refined through alternating manifold projection and energy minimization. These negative samples are subsequently used to optimize the EBM objective $\mathcal{L}_{EBM}$ and the autoencoder objective $\mathcal{L}_{AE}$. Importantly, the autoencoder is trained directly on replay-buffer samples rather than on the original training data. As a result, it progressively learns the structure of the regions explored by Langevin dynamics and becomes an adaptive projection operator that continuously improves replay-buffer quality. The replay buffer is updated using a first-in-first-out strategy and therefore acts as a memory mechanism that stores previously generated low-energy samples, substantially improving the efficiency of Langevin sampling. The complete training procedure is summarized in Algorithm~\ref{algo:joint_training}. Unlike conventional EBM training methods, the proposed framework does not rely exclusively on iterative energy minimization. Instead, each sampling cycle alternates energy minimization and manifold projection. This cooperative interaction allows the autoencoder to regularize Langevin trajectories while enabling the EBM to further refine manifold-projected samples toward lower-energy and more realistic configurations. 

\subsection{Interpretation} 

The proposed framework can be interpreted as the combination of two complementary mechanisms. The autoencoder acts as a manifold projection operator that maps samples toward plausible regions of the data distribution, whereas the EBM acts as an energy-minimization mechanism that further refines these samples according to the learned energy landscape. This interpretation is directly supported by the experimental results presented in Section~\ref{section:experimental}. The jointly trained autoencoder provides the majority of the improvements in both generation and reconstruction quality, while the final EBM refinement consistently reduces sample energy and improves fidelity. Together, these two components enable the proposed framework to simultaneously achieve high-quality generation, robust image reconstruction, and strong out-of-distribution detection performance.
 
\begin{algorithm}[ht] \caption{Cooperative EBM-AE Training} \label{algo:joint_training} \textbf{Input:} training distribution $p_D$, replay buffer $B$, Langevin step size $\eta$, number of Langevin steps $K$, number of EBM-AE-EBM cycles $N$ \begin{algorithmic}[1] \State Pretrain the autoencoder on real data \State Initialize replay buffer $B$ \While{not converged} \State Sample positive examples \[ x^{+}\sim p_D \] \State Initialize samples from replay buffer \[ x_0 \sim 0.95\,B + 0.05\,U(-1,1) \] 
\[ x \leftarrow x_0 \]
\For{$c = 1,\ldots,N$} \State Langevin refinement \[ x \leftarrow \text{Langevin}(x,E_\theta,K) \] \State Add Gaussian perturbation \[ \tilde{x} = x+\sigma\epsilon \] \State Manifold projection \[ x \leftarrow g_\phi(f_\phi(\tilde{x})) \] \State Energy refinement \[ x \leftarrow \text{Langevin}(x,E_\theta,K) \] \EndFor \State Obtain negative samples \[ x^{-}=x \] 
\State Update EBM using \[ \mathcal{L}_{EBM} = \mathbb{E}\!\left[E_{\theta}(x^{+})\right] - \mathbb{E}\!\left[E_{\theta}(x^{-})\right] + \alpha\,\mathcal{R}_{E} \] 
\State Freeze EBM parameters
\State Add Gaussian perturbation and manifold projection \[ \hat{x}^- = g_{\phi}(f_{\phi}(x^- + \sigma\epsilon)) \]
\State Update AE using \[ \mathcal{L}_{AE} = \|x^- -\hat{x}^- \|_{2}^{2} +  \beta E_{\theta}(\hat{x}^-) \] \State Update replay buffer \[ B \leftarrow B \cup x^{-} \] \EndWhile \end{algorithmic} \end{algorithm}


\section{Experimental Results} 
\label{section:experimental}
\subsection{Experimental Setup and Ablation Study} 

The objective of this experimental study is to quantify the respective contributions of the Energy-Based Model (EBM) and the Autoencoder (AE) within the proposed cooperative framework. 
More specifically, we investigate whether the improvement in generation quality originates primarily from the jointly trained autoencoder, from the energy-based refinement stage, or from the combination of both components. 
The experiments were designed to answer this question through a large-scale ablation study conducted on the MNIST dataset.
The proposed framework jointly trains an EBM and a denoising autoencoder. 
During training, the EBM is optimized using Langevin dynamics with ten MCMC sampling steps ($K=10$) and a step size of $\eta = 0.1$, while Gaussian perturbations with standard deviation $0.05$ are applied to the input images in order to improve the robustness of the autoencoder. 
The replay buffer is updated through a single EBM$\rightarrow$AE$\rightarrow$EBM refinement cycle at each training iteration ($N = 1$). 
During generation, however, ten successive refinement cycles are applied to fully exploit the interaction between manifold projection and energy-based optimization ($N=10$). 
To evaluate robustness and reproducibility, thirteen independent training runs were performed using different random initializations. 
For each run, 10,000 images were generated and evaluated using four complementary metrics: 

\begin{itemize} 
\item Fréchet Inception Distance (FID), used to assess the similarity between generated and real samples; 
\item Entropy of the predicted class distribution, measuring generation diversity; 
\item Unknown Rate, corresponding to the percentage of samples not confidently recognized by a reference classifier; 
\item Mean classification confidence, providing an estimate of sample realism. 
\end{itemize} 

In addition to generative metrics, the learned energy function was evaluated through out-of-distribution (OOD) detection experiments using AUROC, AUPR and FPR95. 
For FID computation, a dedicated MNIST classifier was trained from scratch and subsequently used as a feature extractor, providing a dataset-specific alternative to the conventional Inception-based FID. Three generation strategies were compared: 

\begin{enumerate} 
\item \textbf{EBM$\rightarrow$AE$\rightarrow$EBM}: the complete framework combining manifold projection and energy-based refinement. 
\item \textbf{AE Joint}: the autoencoder jointly trained with the EBM but used independently during generation. 
\item \textbf{Vanilla AE}: a conventional denoising autoencoder trained solely on MNIST images.
\end{enumerate} 

\begin{figure}[ht]
    \centering
    \includegraphics[width=6in]{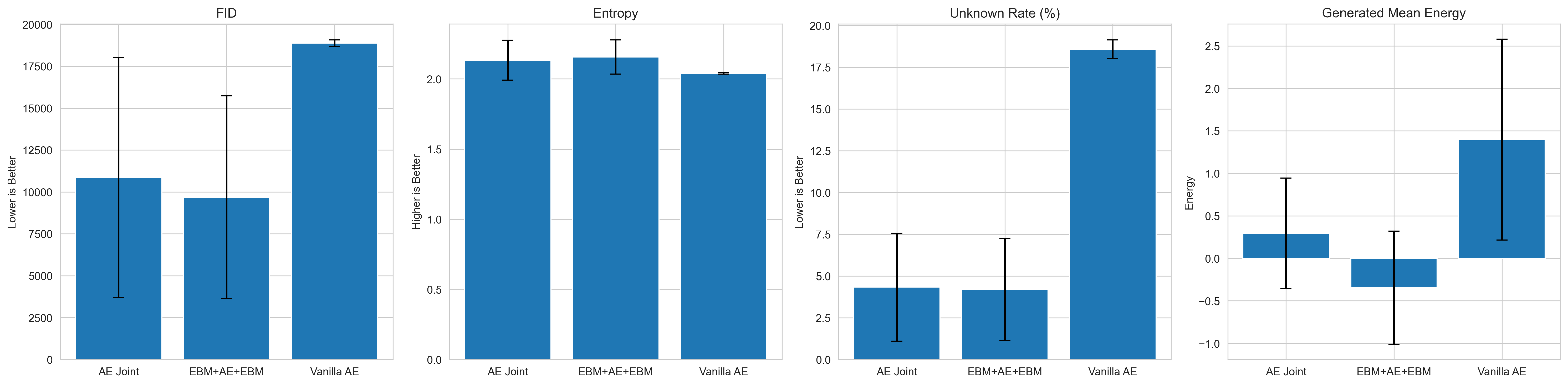}
    \caption{Ablation study conducted over 13 independent experimental runs. The jointly trained models substantially outperform the Vanilla AE, while the complete EBM$\rightarrow$AE$\rightarrow$EBM pipeline provides the best overall trade-off between generation quality, sample validity and diversity.}
    \label{fig:Figure1_AblationStudy}
\end{figure}

The first comparison isolates the contribution of the final energy-refinement stage, while the second evaluates the benefit of the proposed joint-training procedure over a standard autoencoder. 
Since energy-based models do not provide an obvious likelihood measure that can be directly used for model selection, several alternative criteria were investigated throughout the experimental campaign. 
Four checkpoint-selection strategies were evaluated. 
MC1 selects the checkpoint corresponding to the lowest replay-buffer FID, MC2 selects the checkpoint achieving the lowest generated-data FID, MC3 corresponds to the largest energy gap between training samples and replay-buffer samples, and MC4 denotes the final checkpoint obtained at the end of training.
This protocol resulted in a total of 52 evaluated checkpoints for each joint-training configuration, corresponding to thirteen independent runs and four model-selection criteria. 
The Vanilla AE baseline was trained once per run and evaluated using the same generation protocol. 
Figure~\ref{fig:Figure1_AblationStudy} summarizes the complete experimental campaign, while the best observed performances are reported in Table~\ref{tab:global_comparison}.

\begin{table}[ht] \centering \caption{Comparison of generation strategies.} \label{tab:global_comparison} \begin{tabular}{lcccc} \hline Method & FID $\downarrow$ & Entropy $\uparrow$ & Unknown (\%) $\downarrow$ & Confidence $\uparrow$\\ \hline EBM$\rightarrow$AE$\rightarrow$EBM & \textbf{2834.0} & \textbf{2.2855} & \textbf{2.14} & \textbf{0.9967}\\ AE Joint & 3463.0 & 2.2622 & 2.34 & 0.9958\\ Vanilla AE & 18985.0 & 2.0284 & 19.25 & 0.9542\\ \hline \end{tabular} \end{table}

The ablation study reveals three major findings. 
First, jointly training the autoencoder with the EBM dramatically improves generation quality compared with a conventional autoencoder. 
The best FID decreases from 18\,985 for the Vanilla AE to 3\,463 for the jointly trained AE, while the Unknown Rate decreases from approximately 19\% to less than 3\%. 
These results indicate that the replay-buffer supervision and the energy-based objective significantly improve the learned representation of the data manifold. 

Second, the majority of the performance gain originates from the jointly trained autoencoder itself. Even without the final energy-refinement step, the AE Joint configuration consistently generates samples that are substantially more realistic, diverse and recognizable than those produced by the Vanilla AE. 
This observation suggests that the proposed joint-training strategy allows the autoencoder to learn a significantly better approximation of the underlying MNIST manifold. 

Finally, the complete EBM$\rightarrow$AE$\rightarrow$EBM pipeline further improves generation quality and generates the best overall results. The improvement relative to AE Joint is smaller than the gain obtained through joint training, but remains systematic across the experimental campaign. The complete framework achieves the lowest FID score, the lowest Unknown Rate and the highest confidence values among all evaluated methods. 

\begin{figure}[htbp]
    \centering
    
    \begin{subfigure}[b]{0.30\textwidth}
        \centering
        \includegraphics[width=\textwidth]{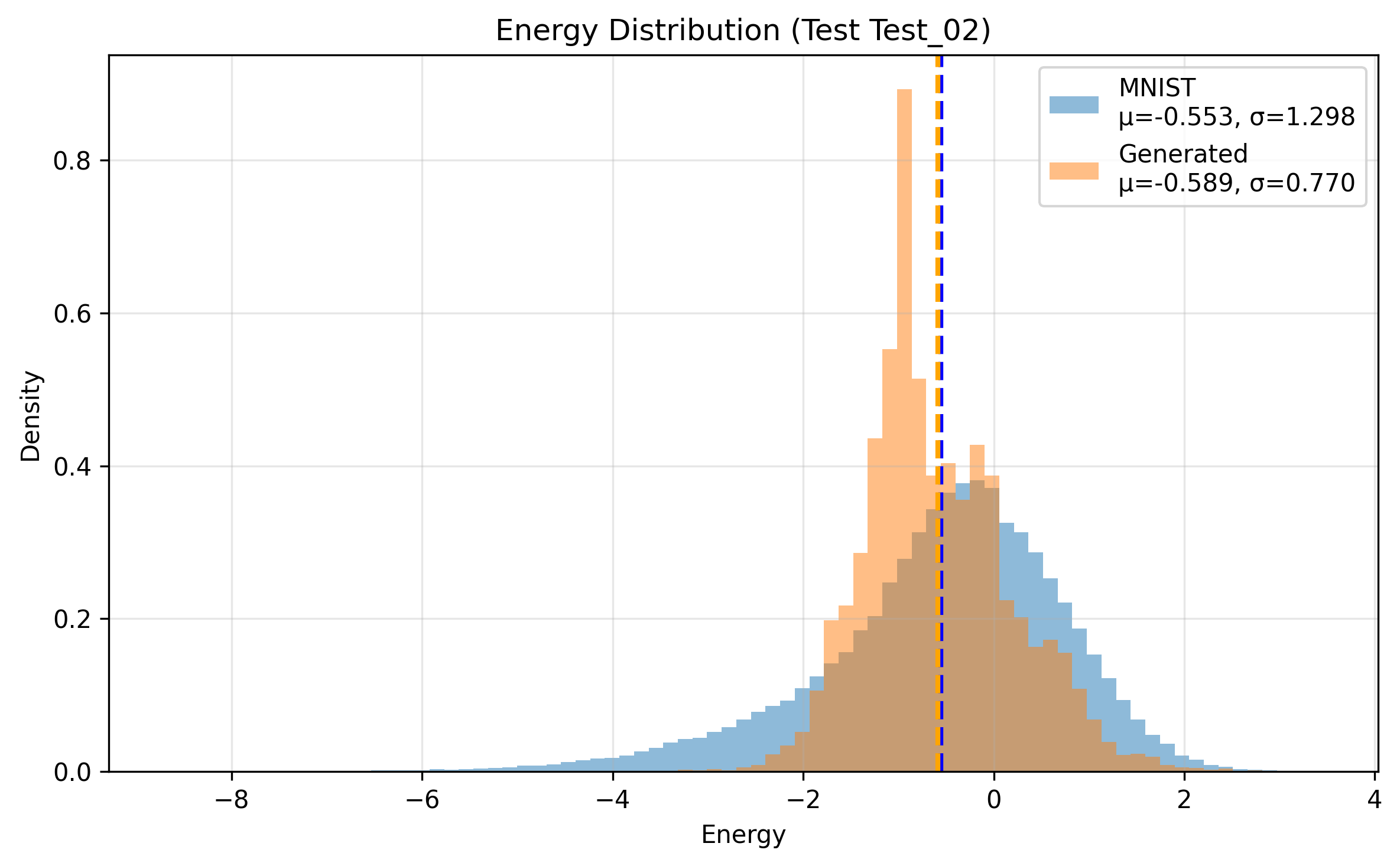}
        \label{fig:3x3-1}
    \end{subfigure}
    \hfill
    \begin{subfigure}[b]{0.30\textwidth}
        \centering
        \includegraphics[width=\textwidth]{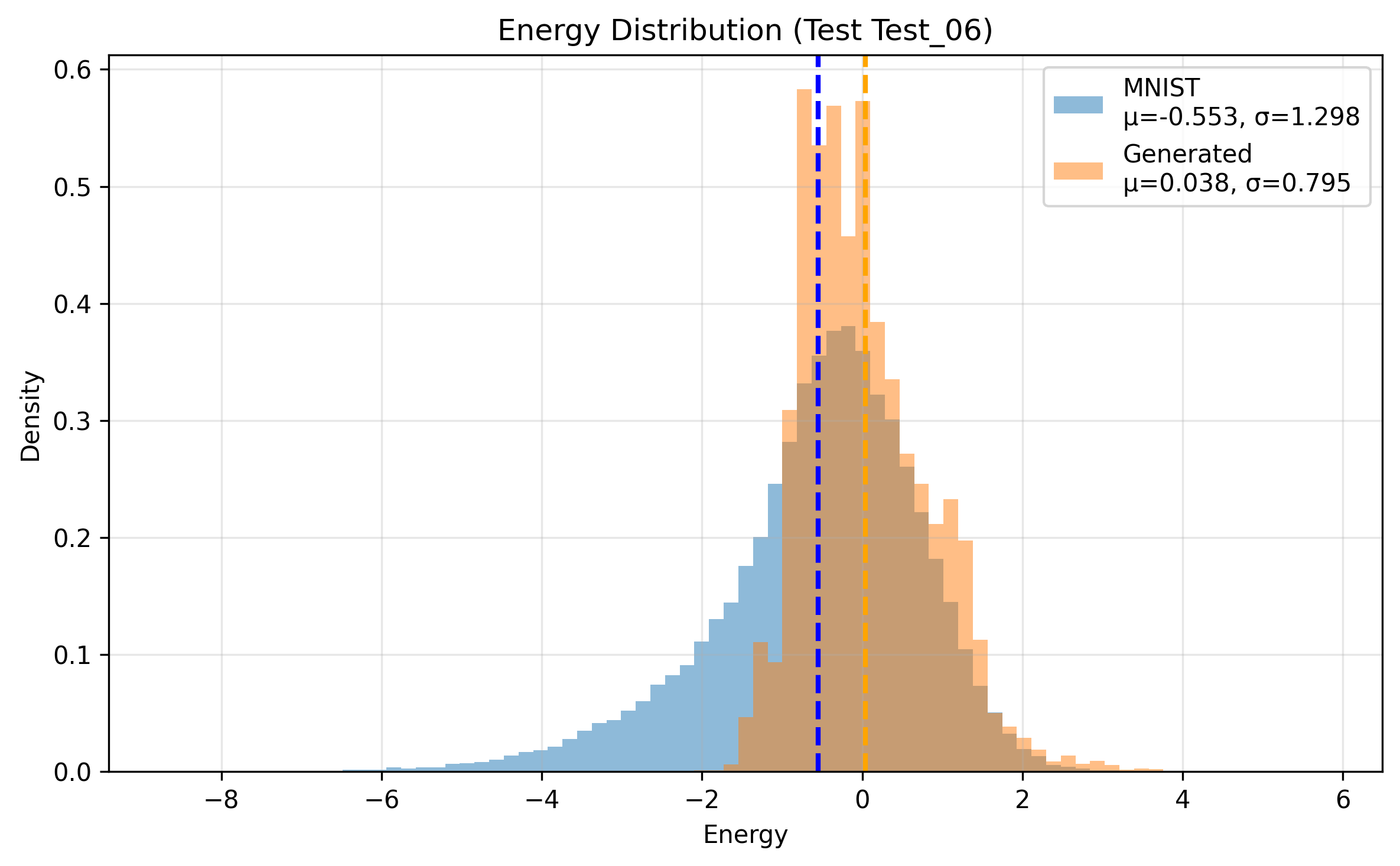}
        \label{fig:3x3-2}
    \end{subfigure}
    \hfill
    \begin{subfigure}[b]{0.30\textwidth}
        \centering
        \includegraphics[width=\textwidth]{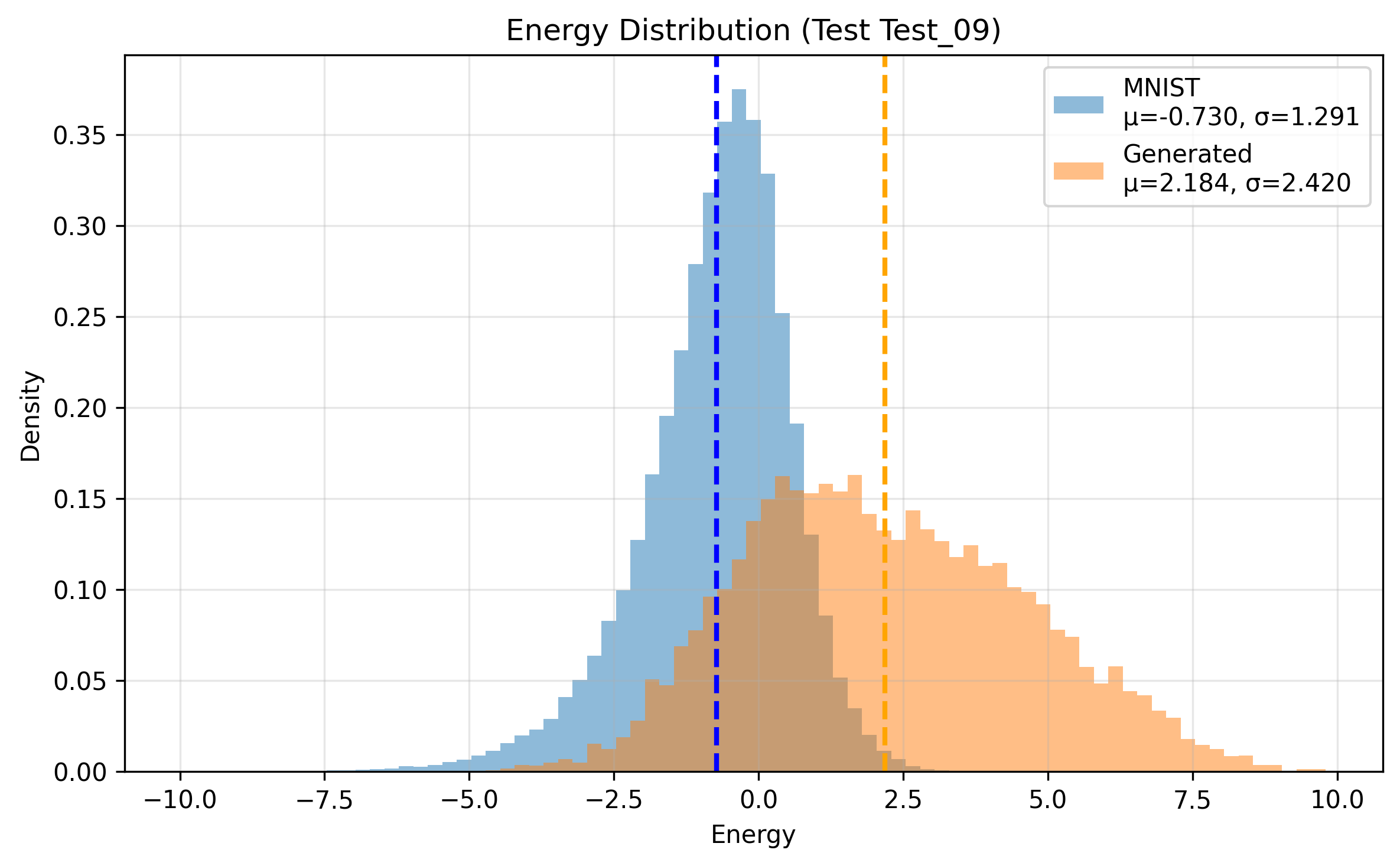}
        \label{fig:3x3-3}
    \end{subfigure}
    
    
    \begin{subfigure}[b]{0.30\textwidth}
        \centering
        \includegraphics[width=\textwidth]{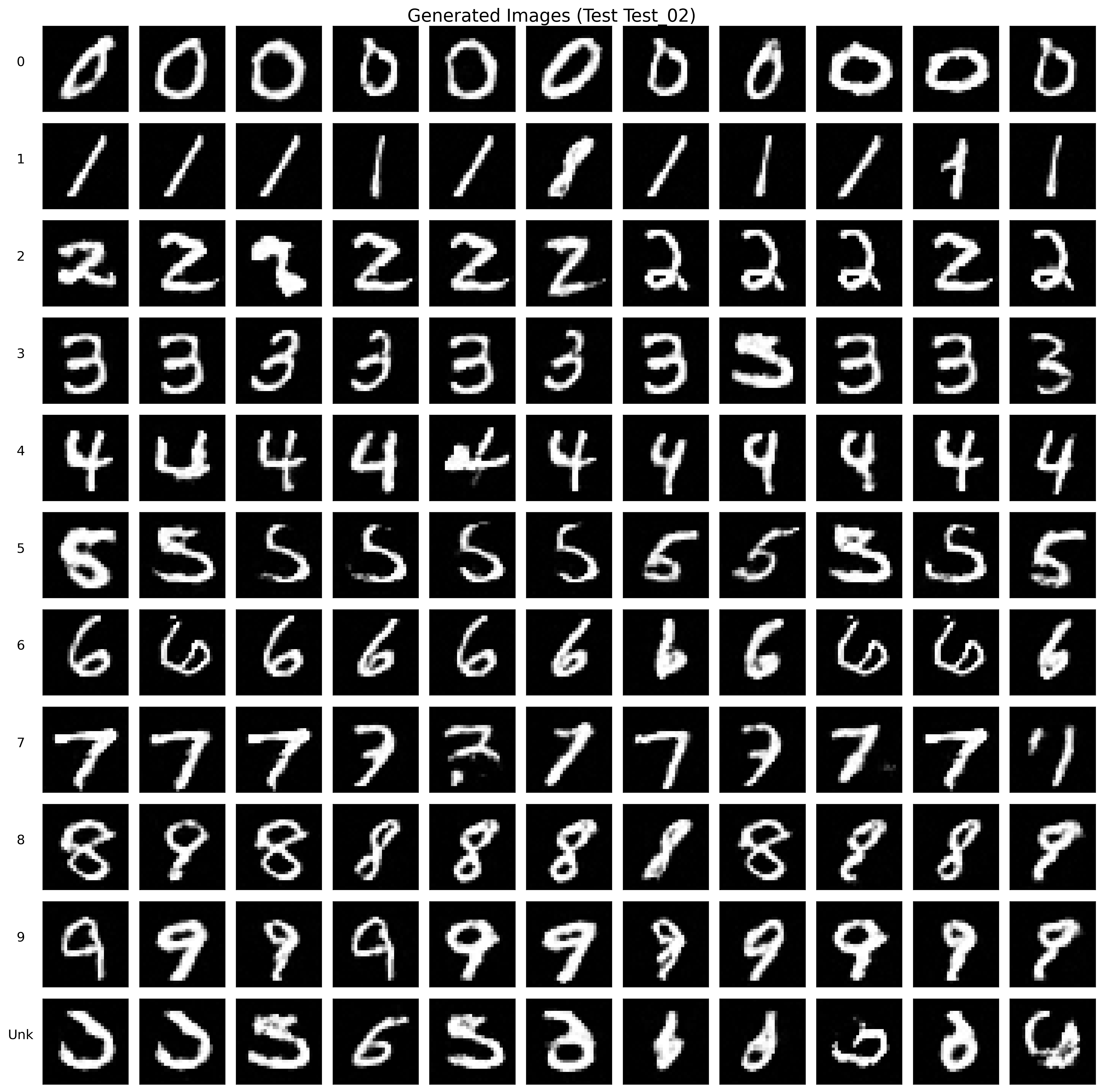}
        \caption{ EBM$\rightarrow$AE$\rightarrow$EBM.}
        \label{fig:imgs_by_class_Test_02}
    \end{subfigure}
    \hfill
    \begin{subfigure}[b]{0.30\textwidth}
        \centering
        \includegraphics[width=\textwidth]{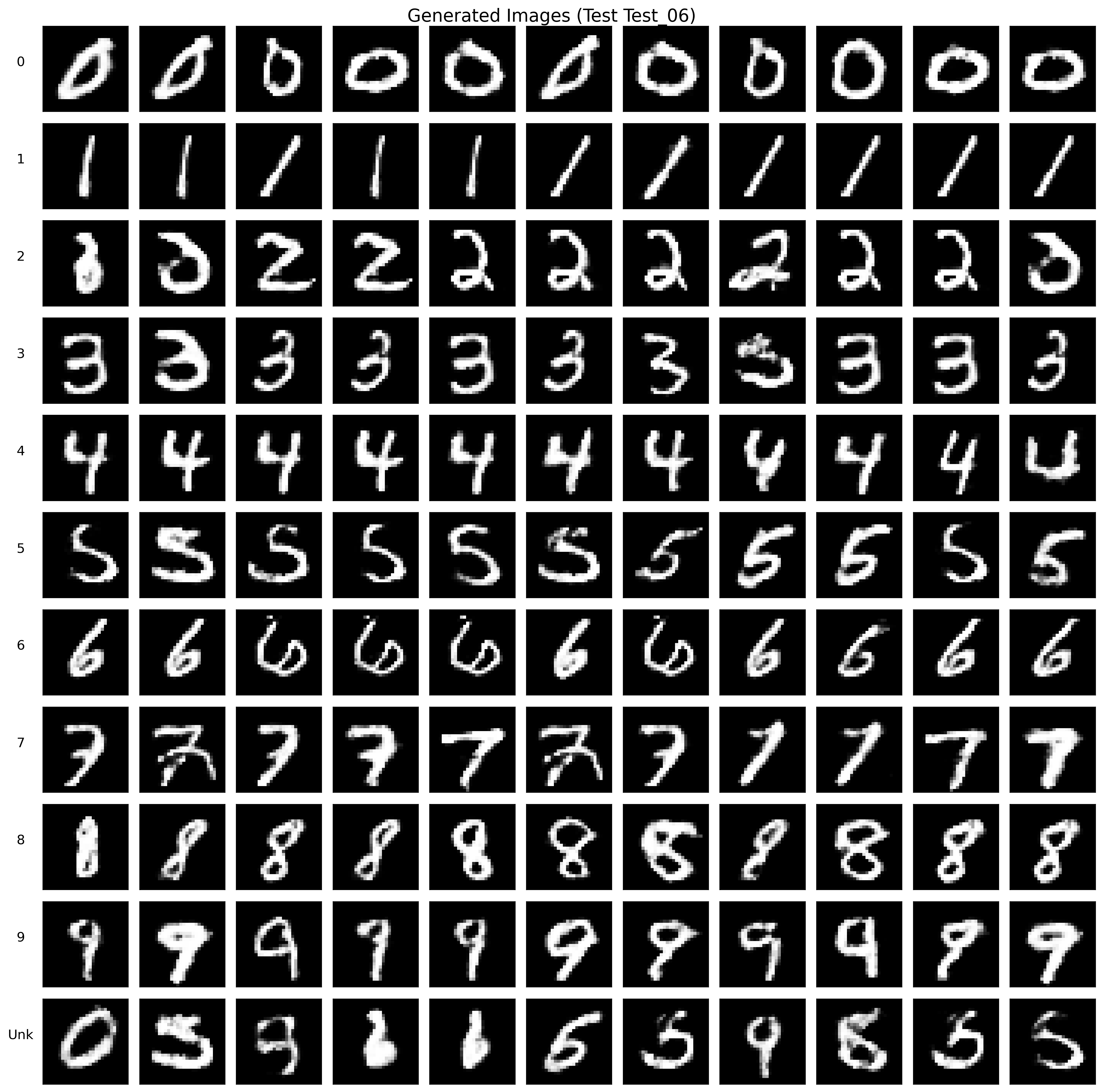}
        \caption{AE Joint.}
        \label{fig:imgs_by_class_Test_06}
    \end{subfigure}
    \hfill
    \begin{subfigure}[b]{0.30\textwidth}
        \centering
        \includegraphics[width=\textwidth]{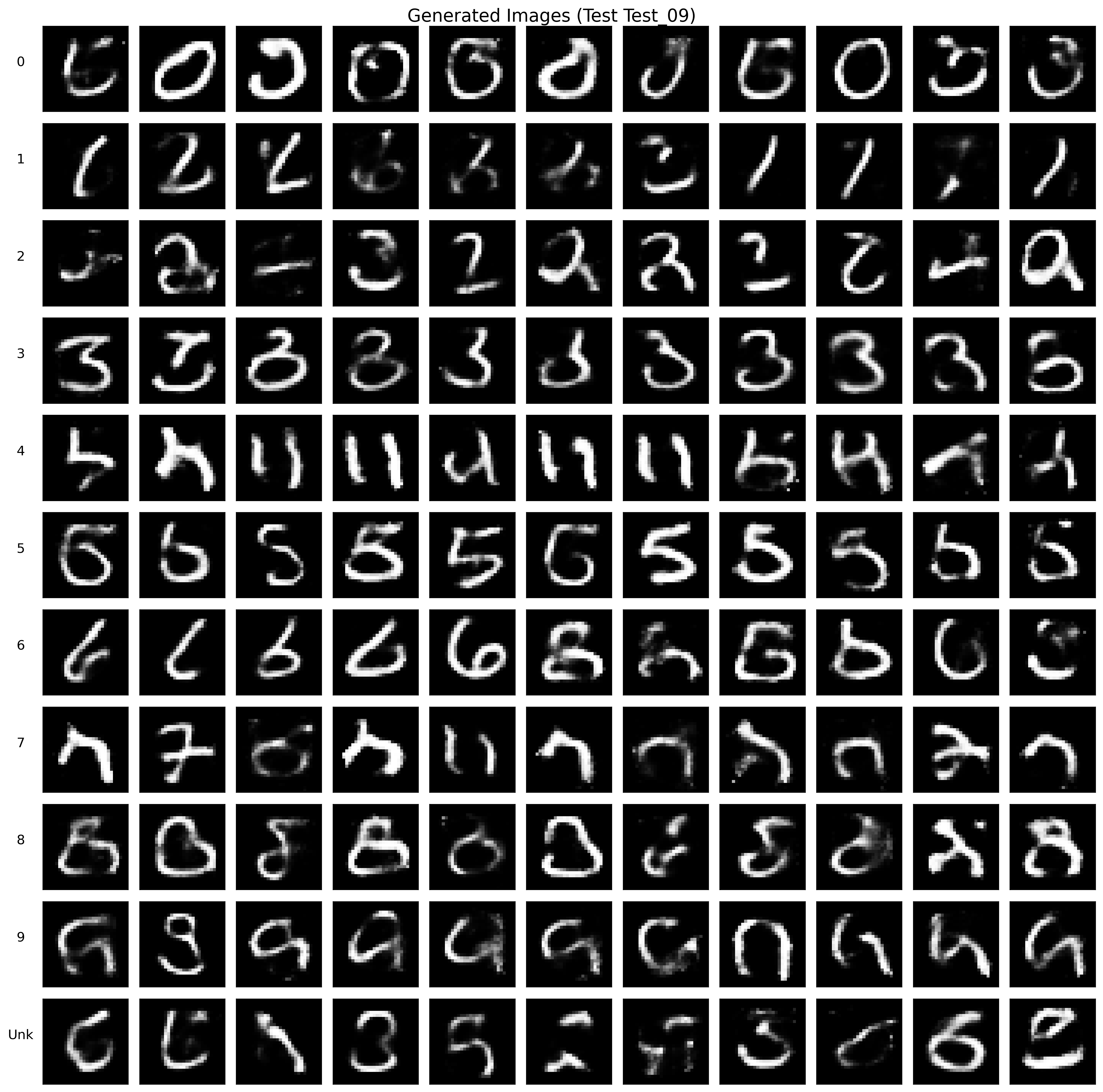}
        \caption{vanilla AE.}
        \label{fig:imgs_by_class_Test_09}
    \end{subfigure}
    
    \vspace{0.3cm} 

    \caption{Qualitative comparison of generated samples and class-wise generation results for the three evaluated methods. The EBM$\rightarrow$AE$\rightarrow$EBM framework produces the most realistic samples while preserving a balanced coverage of digit classes.}
    \label{fig:Qualitative_comparison}
\end{figure}

Figure~\ref{fig:Qualitative_comparison} provides a qualitative comparison of the generated images. The visual examples are consistent with the quantitative results reported in Table~\ref{tab:global_comparison}. Samples generated by the Vanilla AE frequently exhibit blurred digit shapes, artifacts, and reduced class diversity. In contrast, both jointly trained approaches produce sharper and more realistic digits. The complete EBM$\rightarrow$AE$\rightarrow$EBM pipeline generates the most visually coherent samples while maintaining balanced coverage across digit classes. The energy histograms displayed in Figure~\ref{fig:Qualitative_comparison} further support this interpretation. Samples produced by the proposed framework are concentrated in lower-energy regions of the learned landscape, whereas Vanilla AE generations tend to occupy higher-energy regions. This observation suggests that the energy-based refinement stage effectively guides generated samples toward regions that are more consistent with the learned data distribution. 

Taken together, the quantitative results of Figure~\ref{fig:Figure1_AblationStudy} and Table~\ref{tab:global_comparison}, as well as the qualitative examples presented in Figure~\ref{fig:Qualitative_comparison}, support the interpretation that the jointly trained autoencoder acts primarily as a manifold projection operator, while the EBM serves as an energy-minimization mechanism that further refines samples toward low-energy and high-density regions of the learned manifold.

\subsection{Influence of Model Selection Criteria} 

The previous ablation study demonstrated the respective contributions of the jointly trained autoencoder and the final energy-based refinement stage. 
However, the quality of the generated samples also depends strongly on the checkpoint selection strategy. 
 
\begin{figure}[htbp] \centering 
\begin{subfigure}[b]{0.23\textwidth} 
\includegraphics[width=\textwidth] {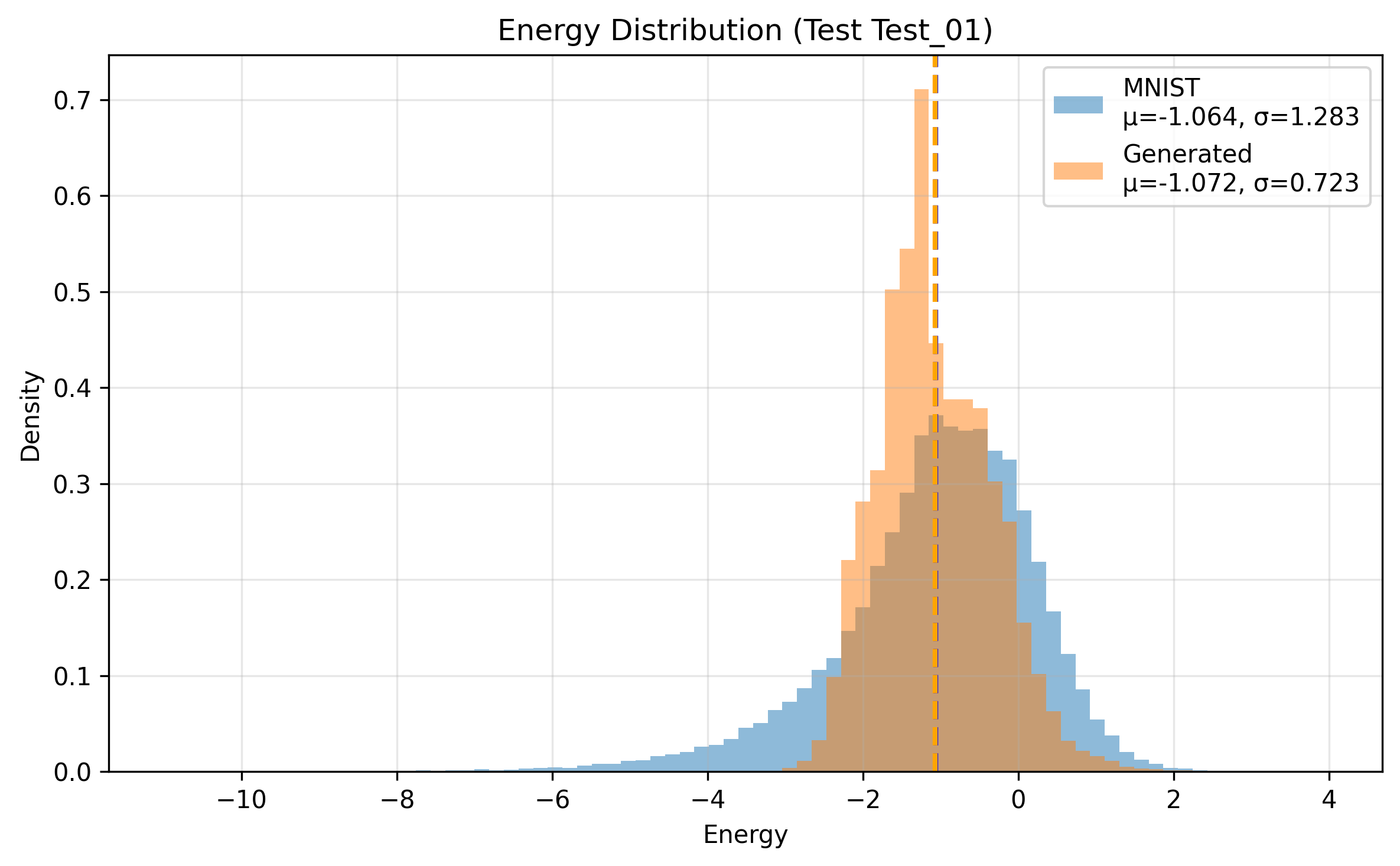} 
\end{subfigure} 
\hfill 
\begin{subfigure}[b]{0.23\textwidth} 
\includegraphics[width=\textwidth] {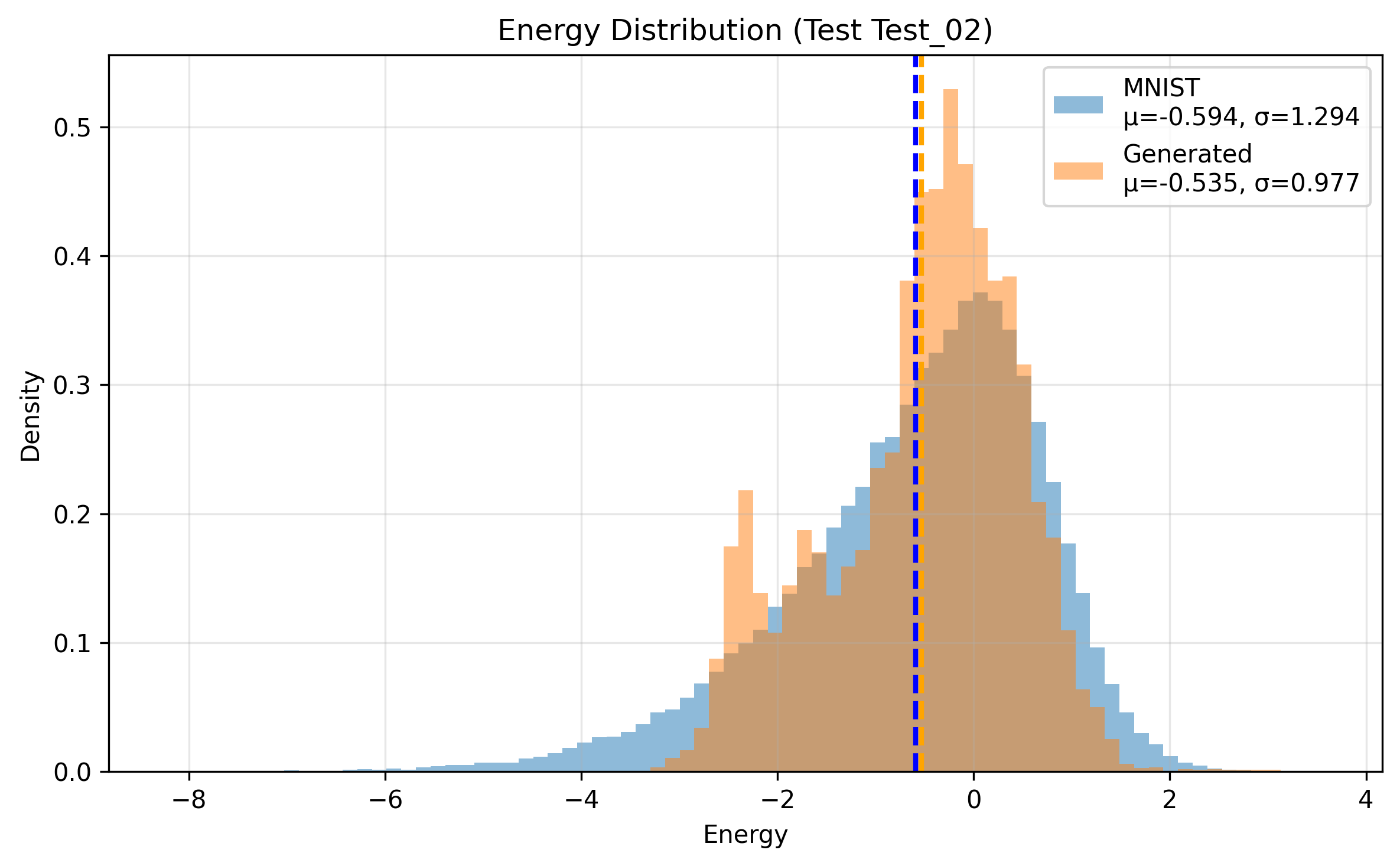} 
\end{subfigure} 
\hfill 
\begin{subfigure}[b]{0.23\textwidth} 
\includegraphics[width=\textwidth] {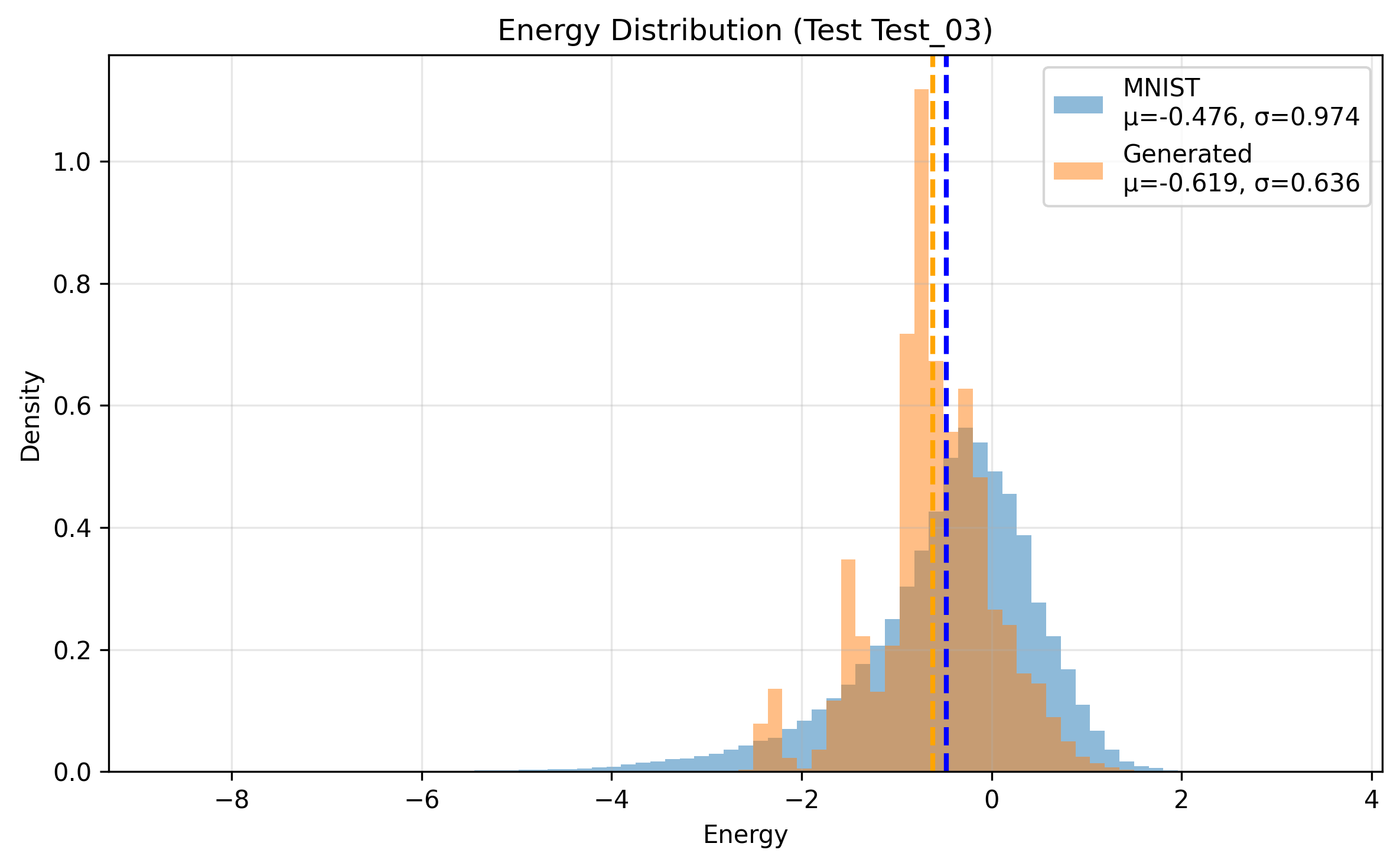} 
\end{subfigure} 
\hfill 
\begin{subfigure}[b]{0.23\textwidth} 
\includegraphics[width=\textwidth] {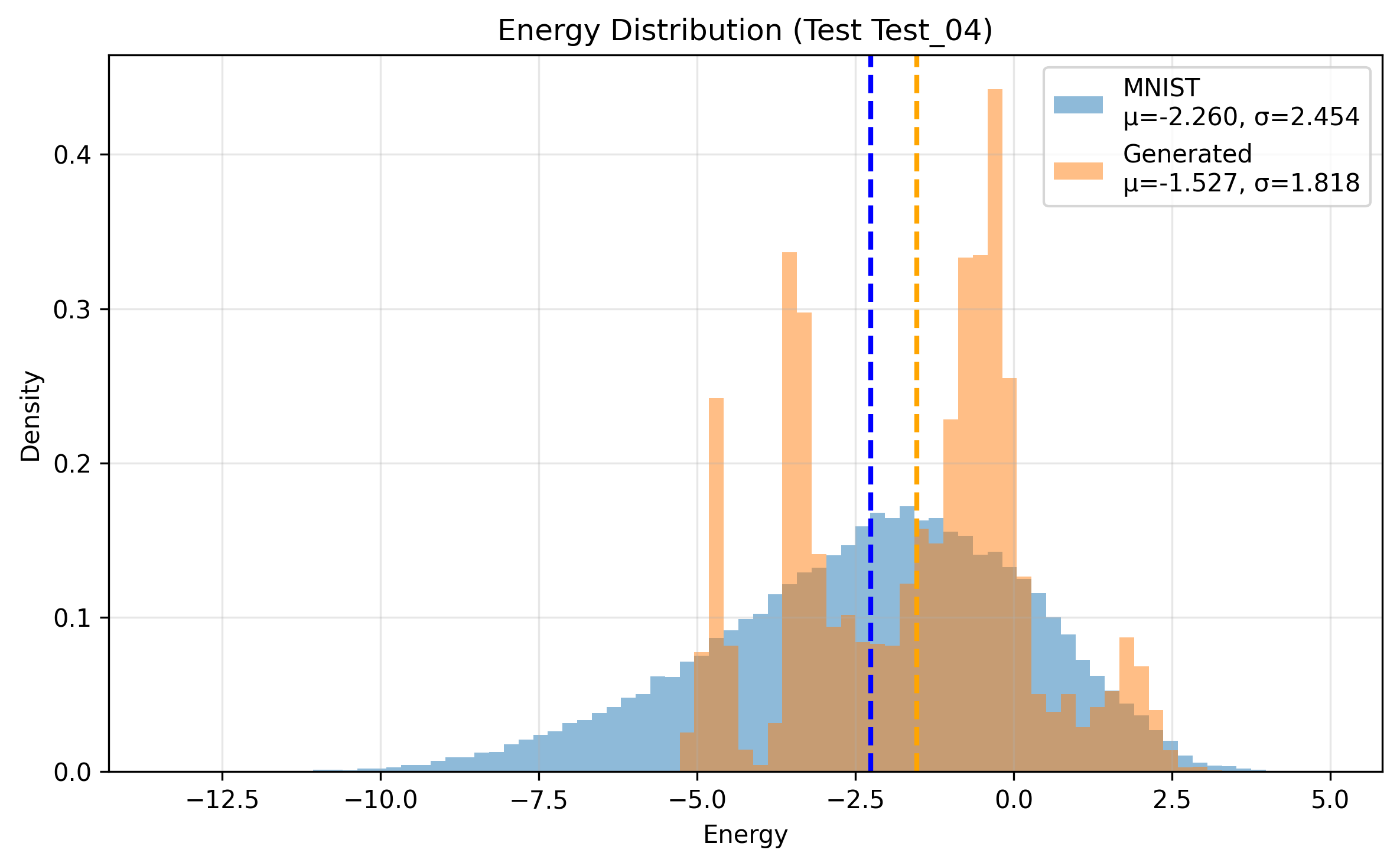} \end{subfigure} 
\vspace{0.3cm} 
\begin{subfigure}[b]{0.23\textwidth} 
\includegraphics[width=\textwidth] {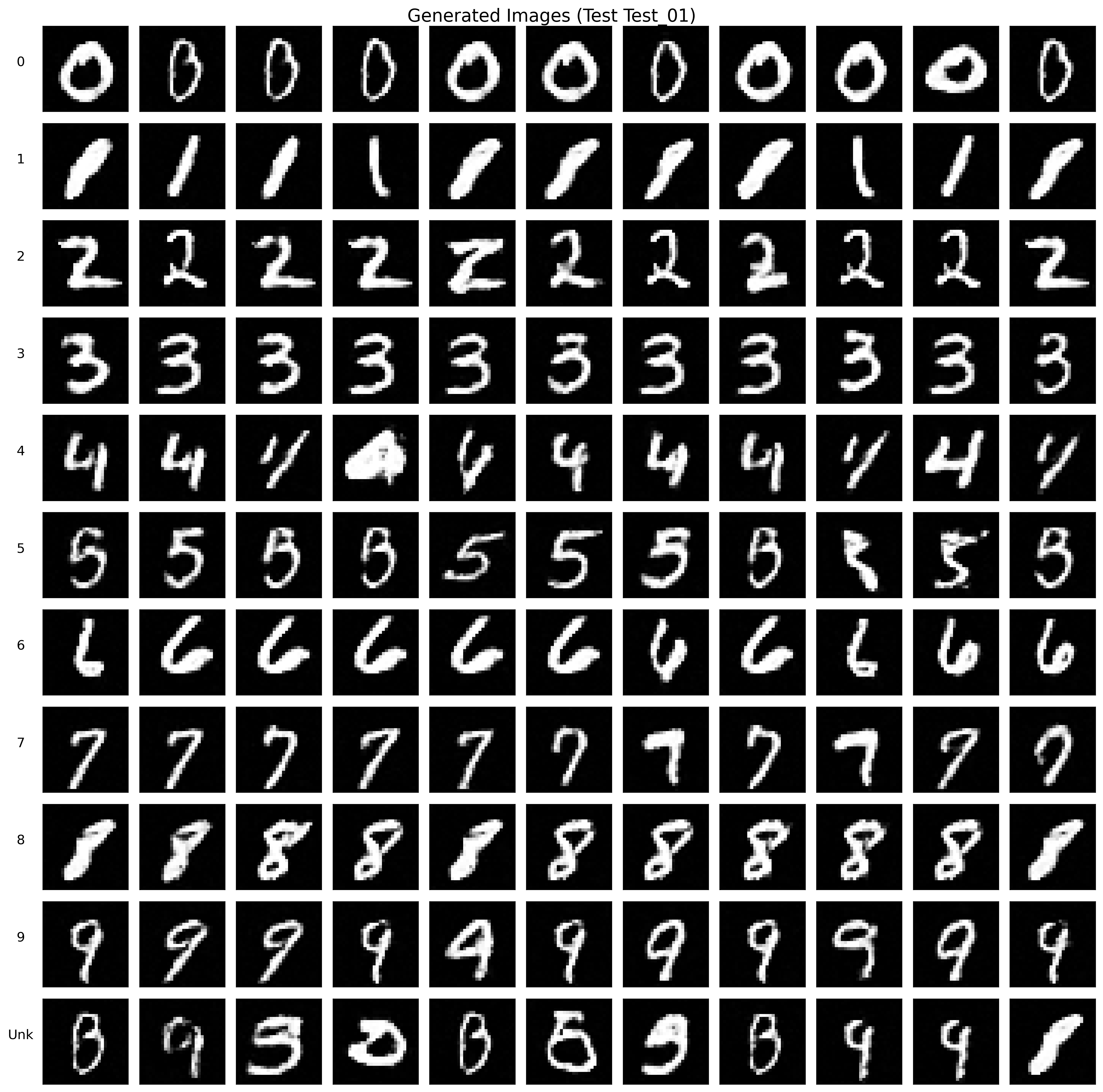} 
\caption{}
\end{subfigure} 
\hfill 
\begin{subfigure}[b]{0.23\textwidth} 
\includegraphics[width=\textwidth] {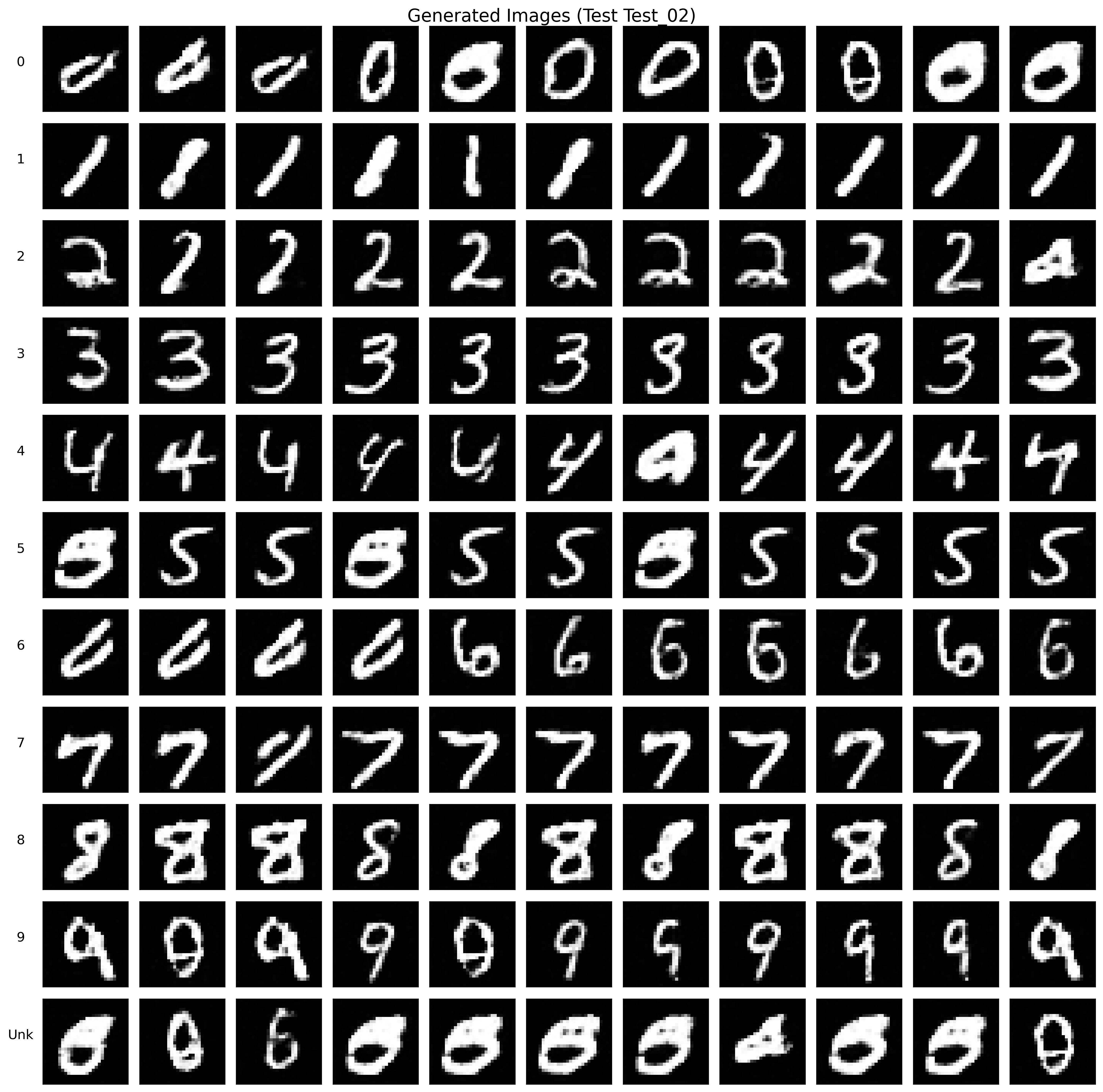} 
\caption{}
\end{subfigure} 
\hfill 
\begin{subfigure}[b]{0.23\textwidth} 
\includegraphics[width=\textwidth] {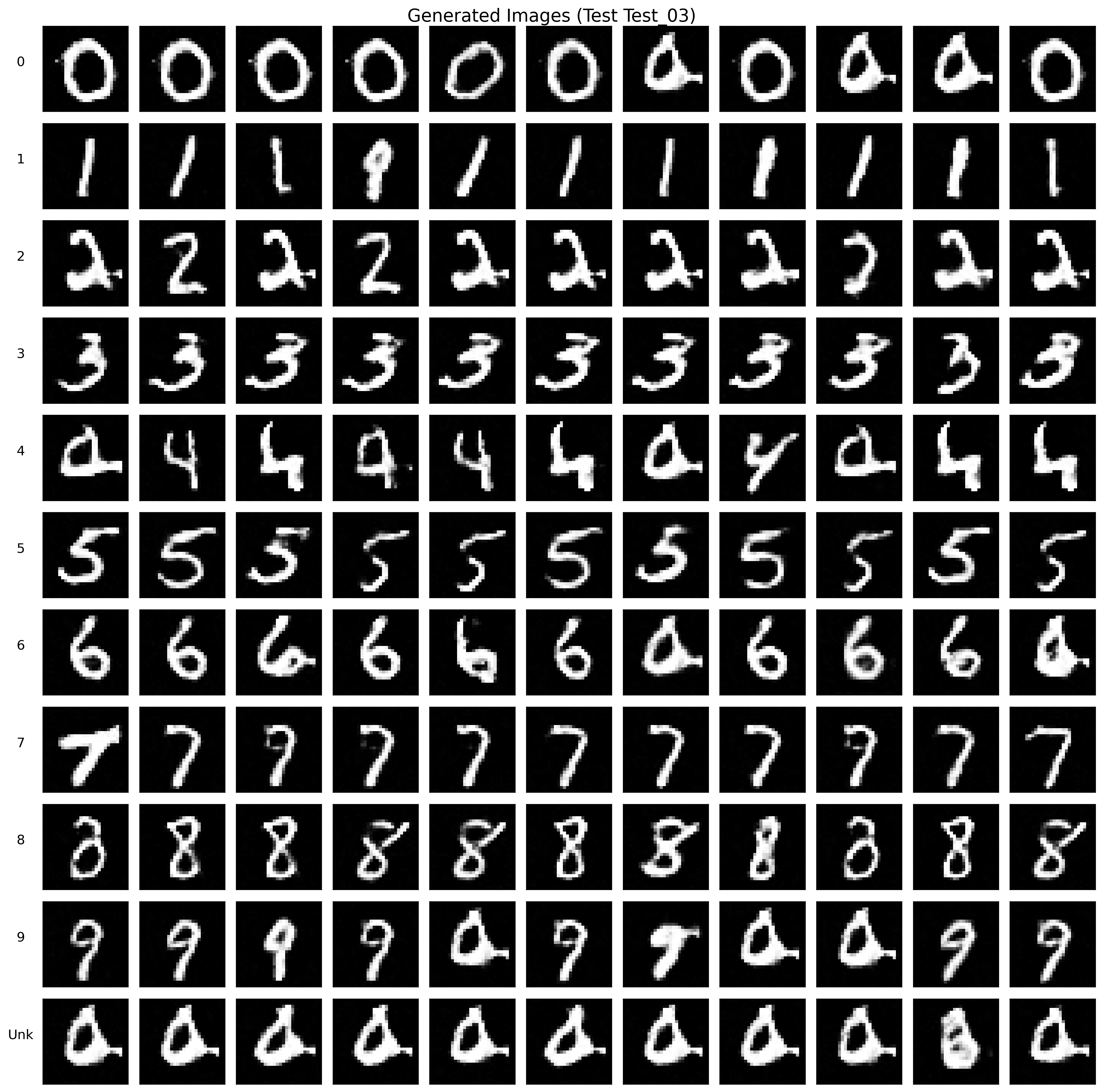} 
\caption{}
\end{subfigure} 
\hfill 
\begin{subfigure}[b]{0.23\textwidth} 
\includegraphics[width=\textwidth] {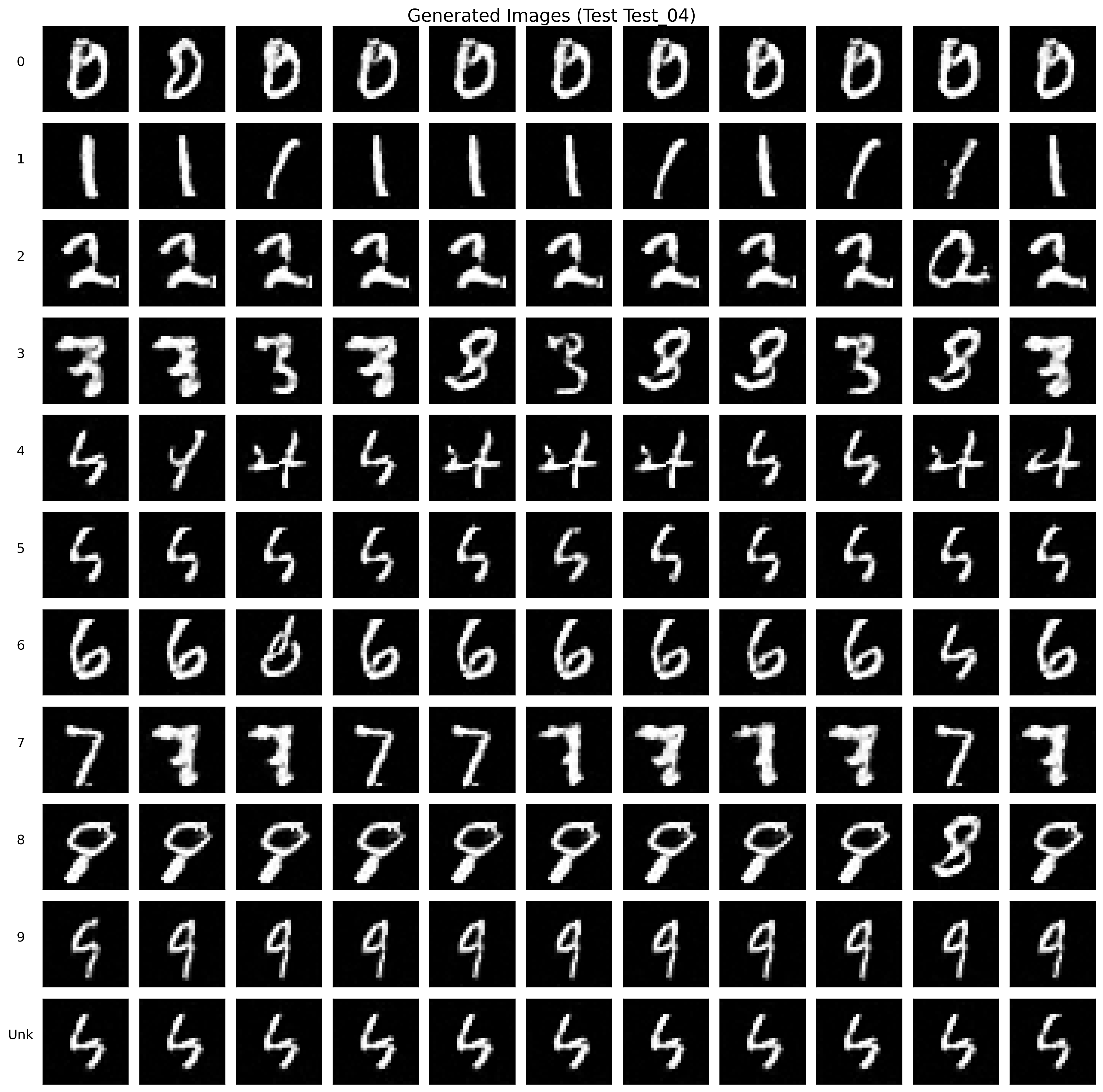} 
\caption{}
\end{subfigure} 
\caption{ Influence of checkpoint selection. Top row: energy histogram comparison. Bottom row: class-wise generated samples. (a) Best replay-buffer FID (MC1), (b) Best generated-data FID (MC2), (c) Best energy-gap checkpoint (MC3), (d) Final checkpoint (MC4). } 
\label{fig:Visual_checkpoint_selection} 
\end{figure}

Table~\ref{tab:modelchoice} summarizes representative results on the influence of checkpoint selection, while Figure~\ref{fig:Visual_checkpoint_selection} provides both qualitative examples and the corresponding energy histograms.

\begin{table}[ht] 
\centering 
\caption{Influence of checkpoint selection.} 
\label{tab:modelchoice} 
\begin{tabular}{lccc} 
\hline 
Selection Criterion & FID & Entropy & Unknown (\%) \\ 
\hline 
MC1 (Best Buffer FID) & 7743 -- 10386 & 2.11 -- 2.25 & 2.3 -- 4.7 \\ 
MC2 (Best Generated FID) & 2834 -- 4589 & 2.26 -- 2.34 & 2.1 -- 4.8 \\ 
MC3 (Best Energy Gap) & 8790 -- 20593 & 1.97 -- 2.23 & 2.2 -- 11.2 \\ 
MC4 (Last Checkpoint) & 5068 -- 16161 & 2.02 -- 2.24 & 1.9 -- 19.0 \\ 
\hline 
\end{tabular} 
\end{table}

The results reveal substantial differences between the four selection strategies. The most important observation is that MC2 consistently achieves the best generative performance. 
Across the experimental campaign, checkpoints selected according to the generated-data FID systematically produce lower FID scores, higher class-distribution entropy, and lower proportions of unrecognized samples than the alternative criteria. 
These results indicate that generated-data FID remains the most reliable indicator of visual sample quality for the proposed framework. 
MC1 generally provides intermediate results. 
Selecting checkpoints according to replay-buffer FID often yields reasonable generation quality, but the generated samples remain consistently inferior to those obtained with MC2. 
This suggests that replay-buffer quality alone does not fully capture the quality of the final samples produced after the complete EBM$\rightarrow$AE$\rightarrow$EBM generation process. 
A particularly interesting behavior is observed for MC3. 
Although maximizing the energy gap is expected to improve the separation between in-distribution and replay-buffer samples, it frequently produces poor generative performance. 
As shown in Table~\ref{tab:modelchoice}, MC3 is associated with the worst FID scores and the lowest entropy values among all evaluated criteria. 
The visual examples in Figure~\ref{fig:Visual_checkpoint_selection} further reveal reduced class diversity and, in several cases, clear signs of mode collapse. 

The energy histograms provide additional insight into this phenomenon. Checkpoints selected using MC3 tend to generate samples concentrated within very narrow regions of the energy landscape. 
While this behavior increases the separation between low-energy and high-energy samples, it also reduces the diversity of generated digits. 
Consequently, a large energy gap should not be interpreted as a direct indicator of generation quality. 
MC4, corresponding to the final training checkpoint, exhibits highly variable behavior. 
Depending on the experimental run, the final checkpoint may remain competitive or may suffer from a noticeable degradation in FID and diversity. 
This observation suggests that the optimal generative model is often reached before the end of training, highlighting the importance of an explicit checkpoint-selection strategy. 
Interestingly, the criterion that maximizes the energy gap (MC3) is precisely the one most closely related to the separation of in-distribution and out-of-distribution samples. This observation motivates the OOD analysis presented in the next section.

\begin{figure}[htbp]
    \centering
\begin{subfigure}[b]{0.30\textwidth}
    \centering
    \includegraphics[width=\textwidth]{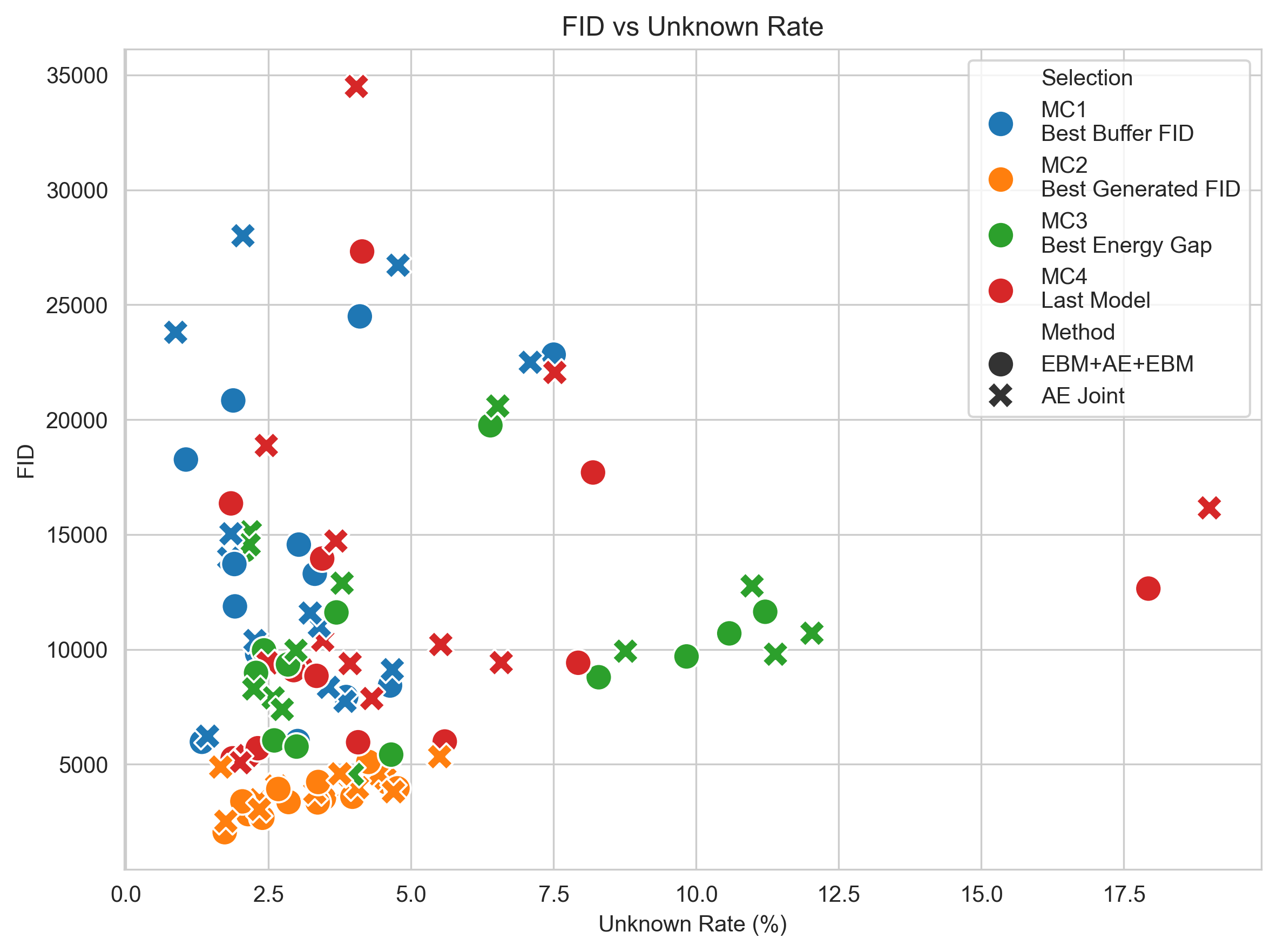}
    \caption{}
    \label{fig:FID_vs_Unknown_bySelection}
\end{subfigure}
\begin{subfigure}[b]{0.30\textwidth}
    \centering
    \includegraphics[width=\textwidth]{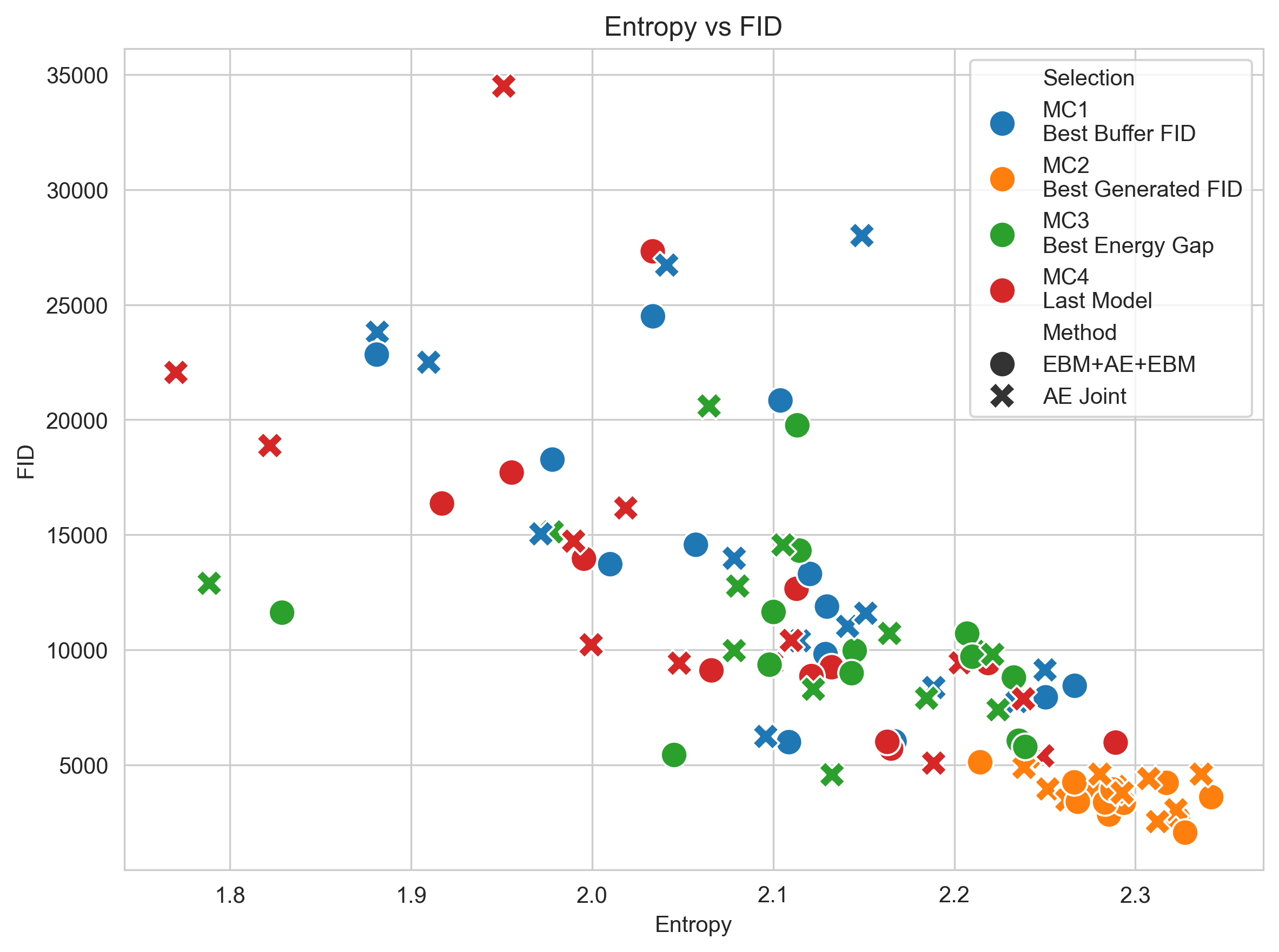}
    \caption{}
    \label{fig:Entropy_vs_FID_bySelection}
\end{subfigure}
\begin{subfigure}[b]{0.30\textwidth}
    \centering 
    \includegraphics[width=\textwidth] {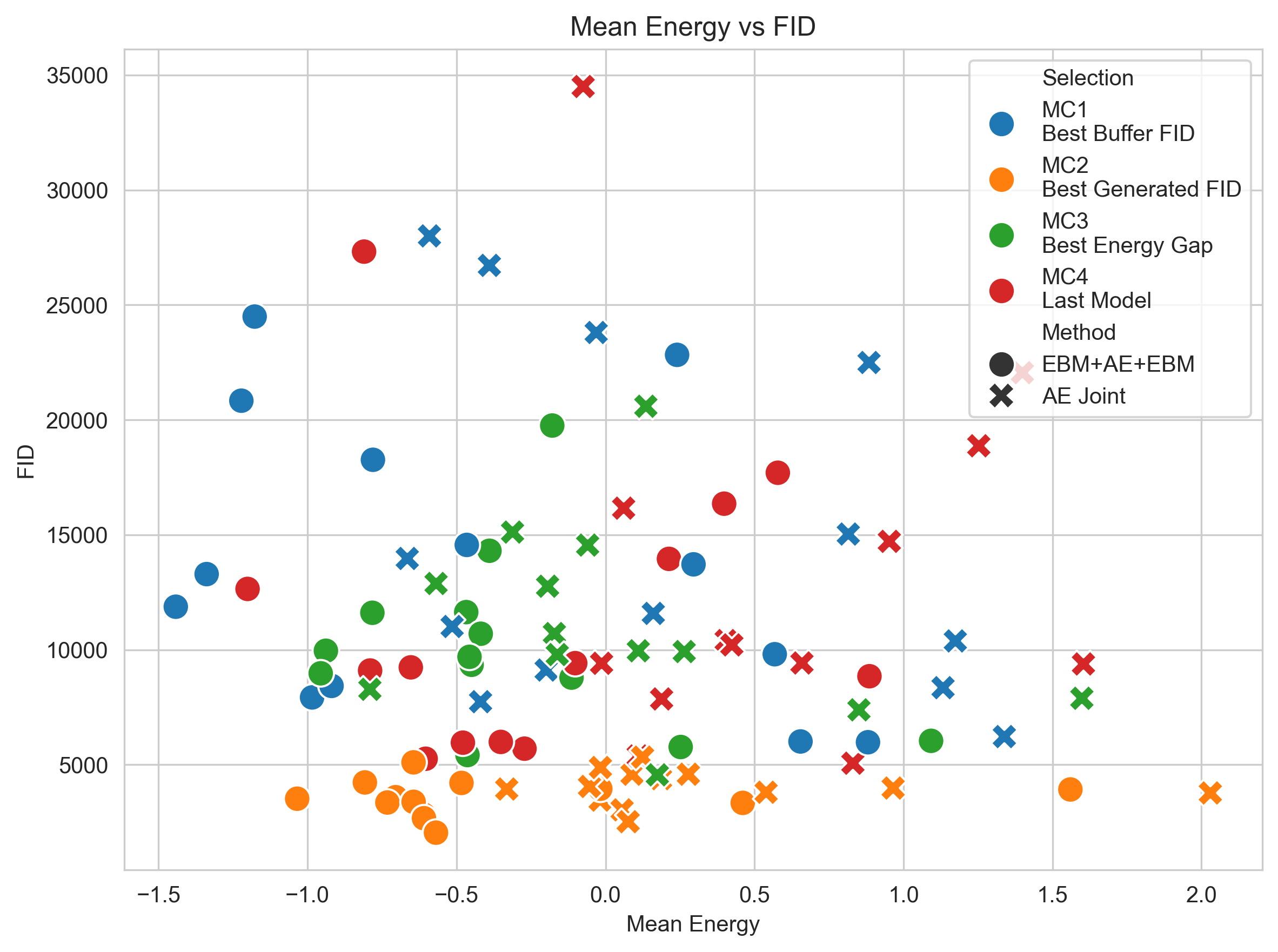} 
    \caption{} 
    \label{fig:MeanEnergy_vs_FID_bySelection} 
\end{subfigure} 
    \caption{Relationship between generation quality and the different evaluation metrics. (a) FID versus Unknown Rate. (b) FID versus class-distribution entropy. (c) FID versus generated-sample energy.}
    \label{fig:metric_relationships}
\end{figure}

Figure~\ref{fig:metric_relationships} further illustrates the relationship between FID and the other evaluation metrics. 
A clear correlation can be observed between FID and both the Unknown Rate and the class-distribution entropy. 
Models achieving lower FID scores generally produce fewer unrecognized samples and exhibit a more balanced class distribution. 
These observations confirm that FID remains a meaningful proxy for both realism and diversity within the proposed framework. 
The third plot relates FID to the average generated-sample energy. 
Although lower-energy samples often correspond to improved generation quality, the relationship is noticeably weaker than for the previous metrics. 
In particular, several checkpoints with comparable energy levels may exhibit significantly different FID scores. 

This observation explains why energy-based selection criteria, such as MC3, do not systematically identify the best generative checkpoints. Taken together, the correlations presented in Figure~\ref{fig:metric_relationships} provide additional justification for the superior performance of MC2. Directly optimizing generated-data FID not only improves visual quality but also leads to favorable values for diversity and sample validity. In contrast, criteria based solely on energy statistics capture only part of the information required to achieve high-quality image generation.
Overall, the results demonstrate that checkpoint selection has a significant impact on the final generation quality. Among all evaluated criteria, MC2 consistently provides the most reliable trade-off between realism, diversity and sample validity. More importantly, these findings reveal that energy-based objectives and generative objectives are not necessarily aligned. A checkpoint exhibiting strong energy separation properties does not automatically correspond to a checkpoint producing visually superior samples. Therefore, energy-based statistics should be viewed as complementary indicators rather than direct surrogates for image-generation quality.

\subsection{Image Inpainting} 

To further evaluate the ability of the proposed framework to exploit partial observations and recover missing information, an image inpainting study was conducted on the MNIST dataset. Unlike unconditional generation, inpainting requires the model to reconstruct plausible samples while remaining consistent with the visible pixels. This task provides a particularly challenging evaluation of the learned manifold and offers additional insight into the respective roles of the autoencoder and the energy-based refinement process. Three reconstruction strategies were compared: 

\begin{enumerate} \item \textbf{EBM$\rightarrow$AE$\rightarrow$EBM}: the complete reconstruction framework combining manifold projection and energy-based refinement; \item \textbf{AE Joint}: the autoencoder jointly trained with the EBM but used independently during reconstruction; \item \textbf{Vanilla AE}: a conventional denoising autoencoder trained solely on MNIST images. 
\end{enumerate} 

For each experiment, a binary mask was applied to the input image and only the visible pixels were kept fixed throughout the reconstruction process. Three masking strategies were considered: 

\begin{itemize} \item Central square masking; \item Vertical strip masking; \item Random pixel masking. 
\end{itemize} 

The masking ratio varied between 20\% and 50\%, allowing the reconstruction difficulty to be progressively increased. 
Table~\ref{tab:inpainting_results} summarizes representative reconstruction results. 
A first observation is that reconstruction quality strongly depends on the amount of missing information. 
For all methods, FID scores increase significantly as the masking ratio becomes larger. 
While relatively low FID values are obtained for moderate masking ratios, reconstruction quality deteriorates rapidly when large portions of the image are removed. 
A second observation concerns the influence of the masking pattern itself. 
Random masks consistently produce lower FID scores than structured masks of comparable size. 
This behavior can be explained by the fact that randomly observed pixels remain distributed across the entire digit, preserving local information in most regions of the image. 
In contrast, central and vertical masks remove contiguous image structures and create more ambiguous reconstruction problems. 
Among the evaluated configurations, vertical masks generally constitute the most challenging scenario. 
The comparison between reconstruction methods reveals trends that closely mirror those observed in the generation experiments. 
Joint EBM-AE training substantially improves reconstruction quality relative to the Vanilla AE baseline. 

\begin{table}[ht] \centering \caption{Inpainting FID performance for different masking configurations.} \label{tab:inpainting_results} \begin{tabular}{lcccc} \toprule Mask Type & Ratio & EBM$\rightarrow$AE$\rightarrow$EBM & AE Joint & Vanilla AE \\ \midrule Center & 0.20 & 710 & 718 & 1288 \\ Center & 0.30 & 2759 & 3021 & 7458 \\ Center & 0.50 & 22230 & 22679 & 29746 \\ Random & 0.20 & 760 & 1048 & \textbf{489} \\ Random & 0.30 & 1940 & 2441 & 1700 \\ Random & 0.50 & 6360 & 6961 & 9127 \\ Vertical & 0.20 & 4300 & 4546 & 10794 \\ Vertical & 0.35 & 15953 & 16042 & 24706 \\ Vertical & 0.50 & 30253 & 30814 & 35682 \\ \bottomrule \end{tabular} 
\end{table} 

For most masking configurations, the jointly trained models achieve significantly lower FID values, indicating that the manifold learned through the cooperative training process provides a much stronger prior than a conventional autoencoder trained exclusively on clean images. 
Interestingly, the majority of the improvement is already obtained with the jointly trained autoencoder itself. 
The AE Joint configuration consistently outperforms the Vanilla AE and remains close to the performance of the complete reconstruction pipeline. 
This observation suggests that joint training enables the autoencoder to learn an accurate approximation of the MNIST manifold and to project corrupted observations toward plausible digit configurations. 
The complete EBM$\rightarrow$AE$\rightarrow$EBM framework nevertheless achieves the best overall reconstruction performance. 
Although the improvement relative to AE Joint remains moderate for easy reconstruction tasks, the benefit of the energy-based refinement stage becomes increasingly visible as the masking ratio grows. 
For highly structured masks and large missing regions, the EBM systematically improves the final reconstruction, indicating that the learned energy function provides useful information beyond the manifold projection performed by the autoencoder alone. To better understand the reconstruction process, we additionally analyzed the evolution of the generated-sample energy. 

Table~\ref{tab:inpainting_energy} reports the average energy values measured before and after reconstruction. A striking observation is that masked images generally exhibit very high energy values. Depending on the masking strategy and masking ratio, the average energy ranges from approximately 3 to more than 13. Furthermore, the energy consistently increases with reconstruction difficulty. For example, the average energy rises from 3.12 for a 20\% central mask to 11.00 for a 50\% central mask. Similar trends can be observed for both random and vertical masks. These results indicate that partially observed images are naturally located far from the low-energy regions associated with valid MNIST digits. In other words, introducing missing regions moves the observations away from the learned data manifold and into higher-energy regions of the model's energy landscape. After reconstruction, a dramatic reduction in energy is observed for all jointly trained approaches. 

\begin{table}[ht] 
\centering 
\caption{Mean generated-sample energy for different masking configurations.} 
\label{tab:inpainting_energy} 
\begin{tabular}{lccccc} 
\toprule 
Mask Type & Ratio & Masked & EBM$\rightarrow$AE$\rightarrow$EBM & AE Joint & Vanilla AE \\ 
\midrule 
Center & 0.20 & 3.12 & \textbf{0.68} & 0.76 & 0.91 \\ 
Center & 0.30 & 6.85 & \textbf{0.91} & 1.16 & 3.41 \\ 
Center & 0.50 & 11.00 & \textbf{1.43} & 2.18 & 7.34 \\ 
Random & 0.20 & 6.86 & 1.53 & 1.84 & \textbf{0.97} \\ 
Random & 0.30 & 8.58 & 1.99 & 2.37 & \textbf{1.46} \\ 
Random & 0.50 & 13.15 & \textbf{2.63} & 3.15 & 3.03 \\ 
Vertical & 0.20 & 8.44 & \textbf{1.12} & 1.50 & 4.17 \\ 
Vertical & 0.35 & 10.36 & \textbf{1.37} & 2.41 & 6.49 \\ 
Vertical & 0.50 & 12.82 & \textbf{1.76} & 2.59 & 7.48 \\ 
\bottomrule 
\end{tabular} 
\end{table}

\begin{figure*}[t] \centering 

\begin{subfigure}[b]{0.30\textwidth} 
\centering 
\includegraphics[width=\textwidth] {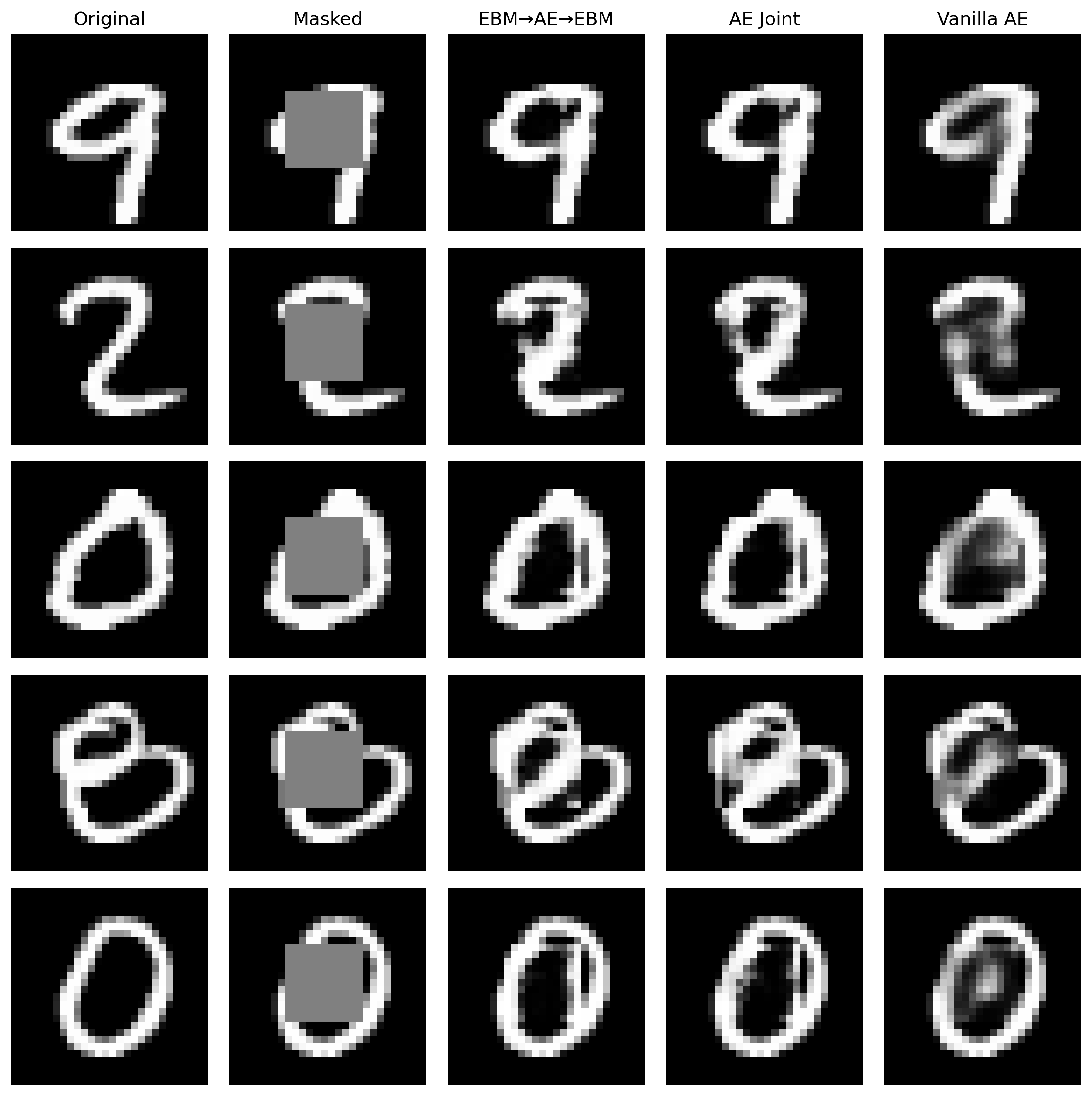} 
\label{fig:center40_reconstruction} 
\end{subfigure} 
\begin{subfigure}[b]{0.30\textwidth} 
\centering 
\includegraphics[width=\textwidth] {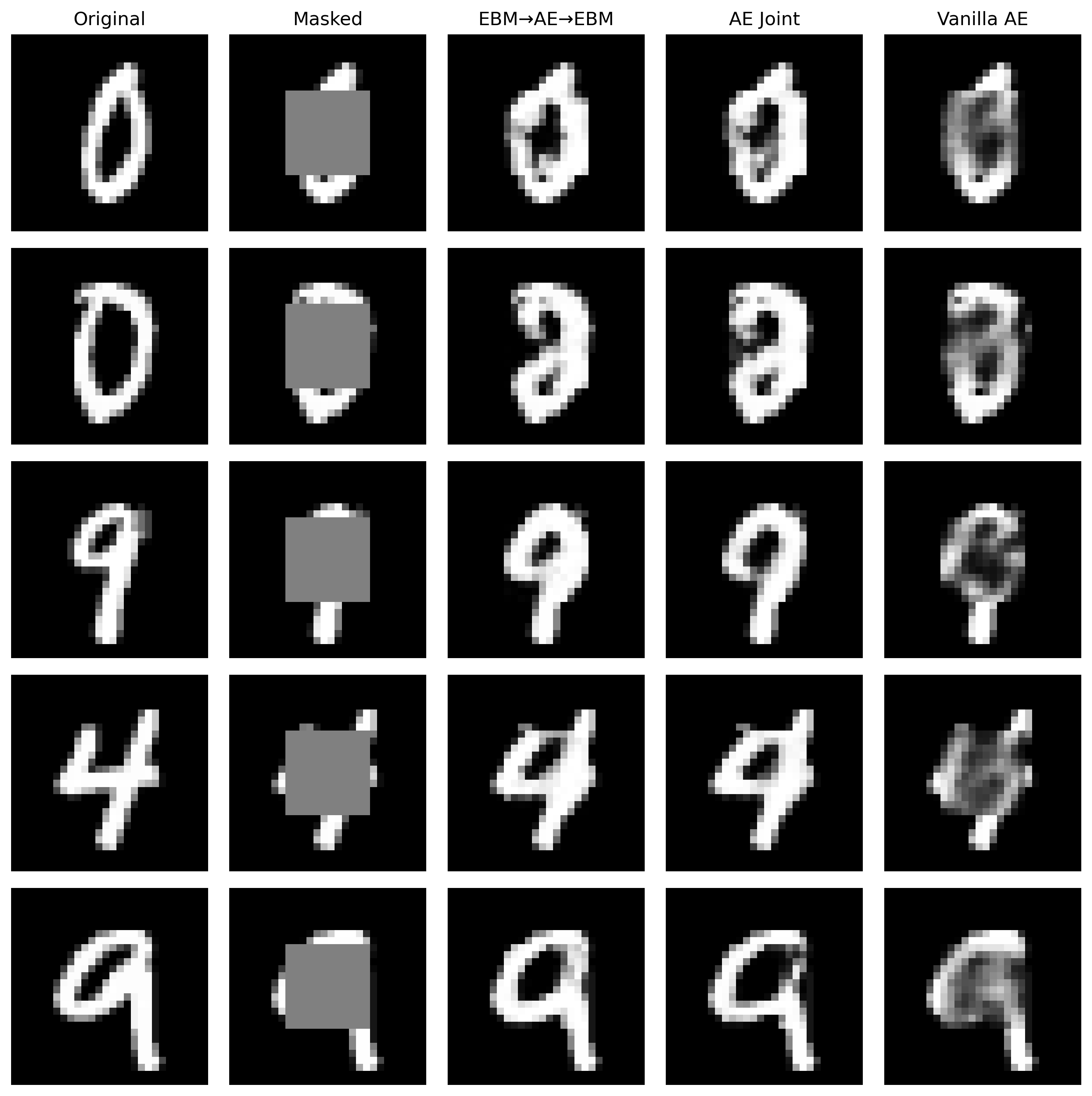} 
\label{fig:center45_reconstruction} 
\end{subfigure} 
\begin{subfigure}[b]{0.30\textwidth} 
\centering 
\includegraphics[width=\textwidth] {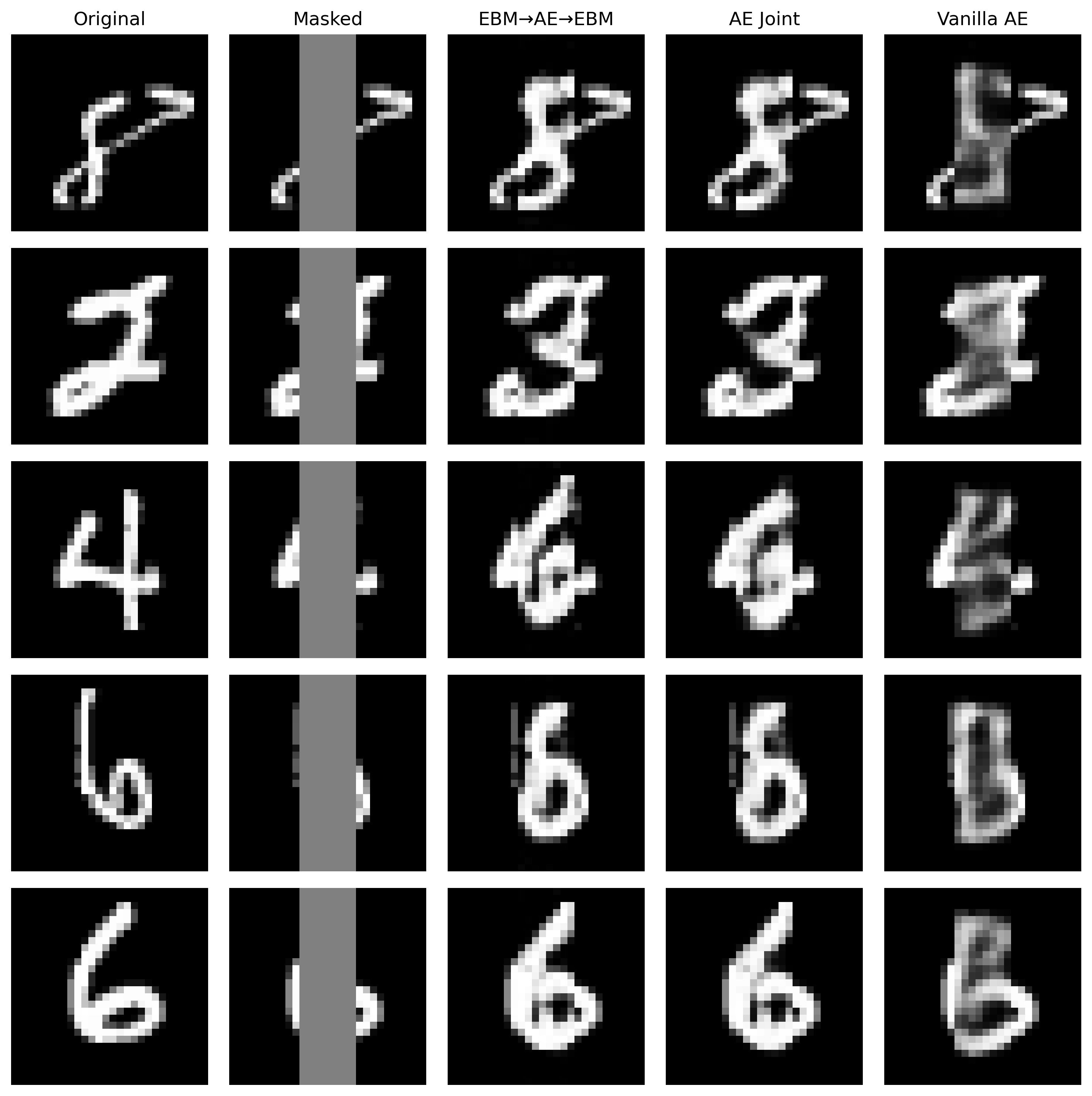} 
\label{fig:center45_reconstruction} 
\end{subfigure} 
\vspace{0.4cm} 
\begin{subfigure}[b]{0.30\textwidth} 
\centering 
\includegraphics[width=\textwidth] {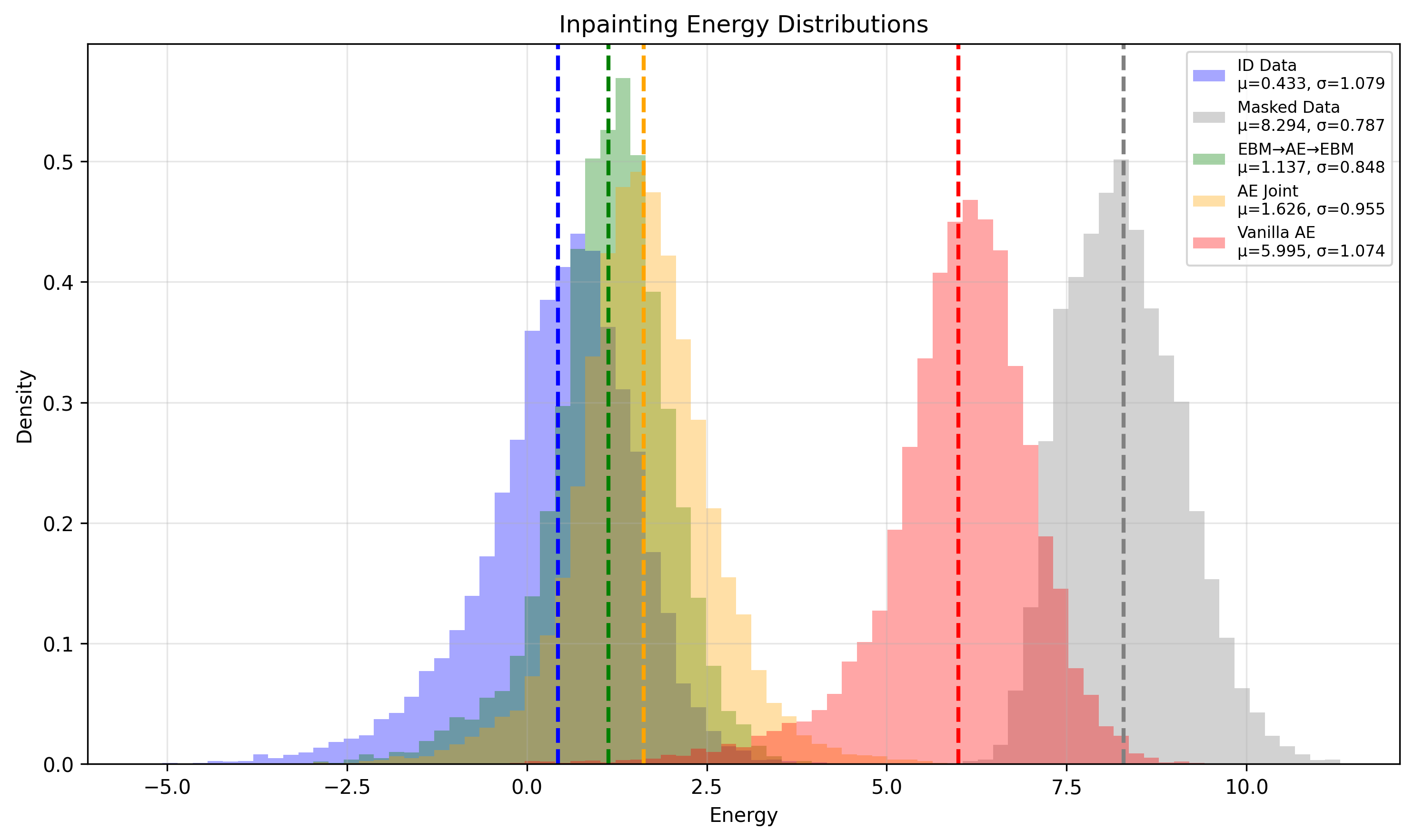} 
\caption{} 
\label{fig:center40_energy} 
\end{subfigure} 
\begin{subfigure}[b]{0.30\textwidth} 
\centering 
\includegraphics[width=\textwidth] {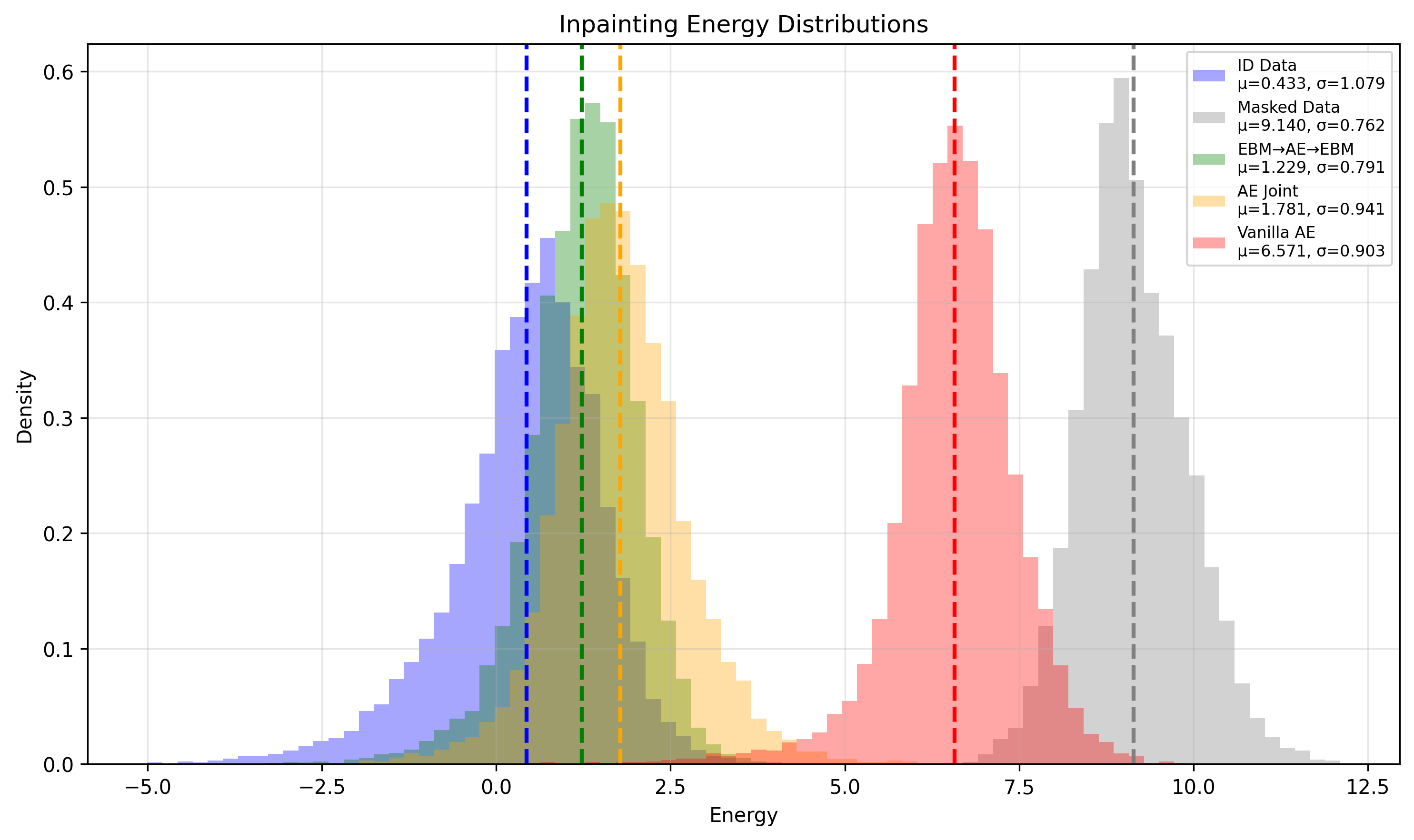} 
\caption{} 
\label{fig:center45_energy} 
\end{subfigure} 
\begin{subfigure}[b]{0.30\textwidth} 
\centering 
\includegraphics[width=\textwidth] {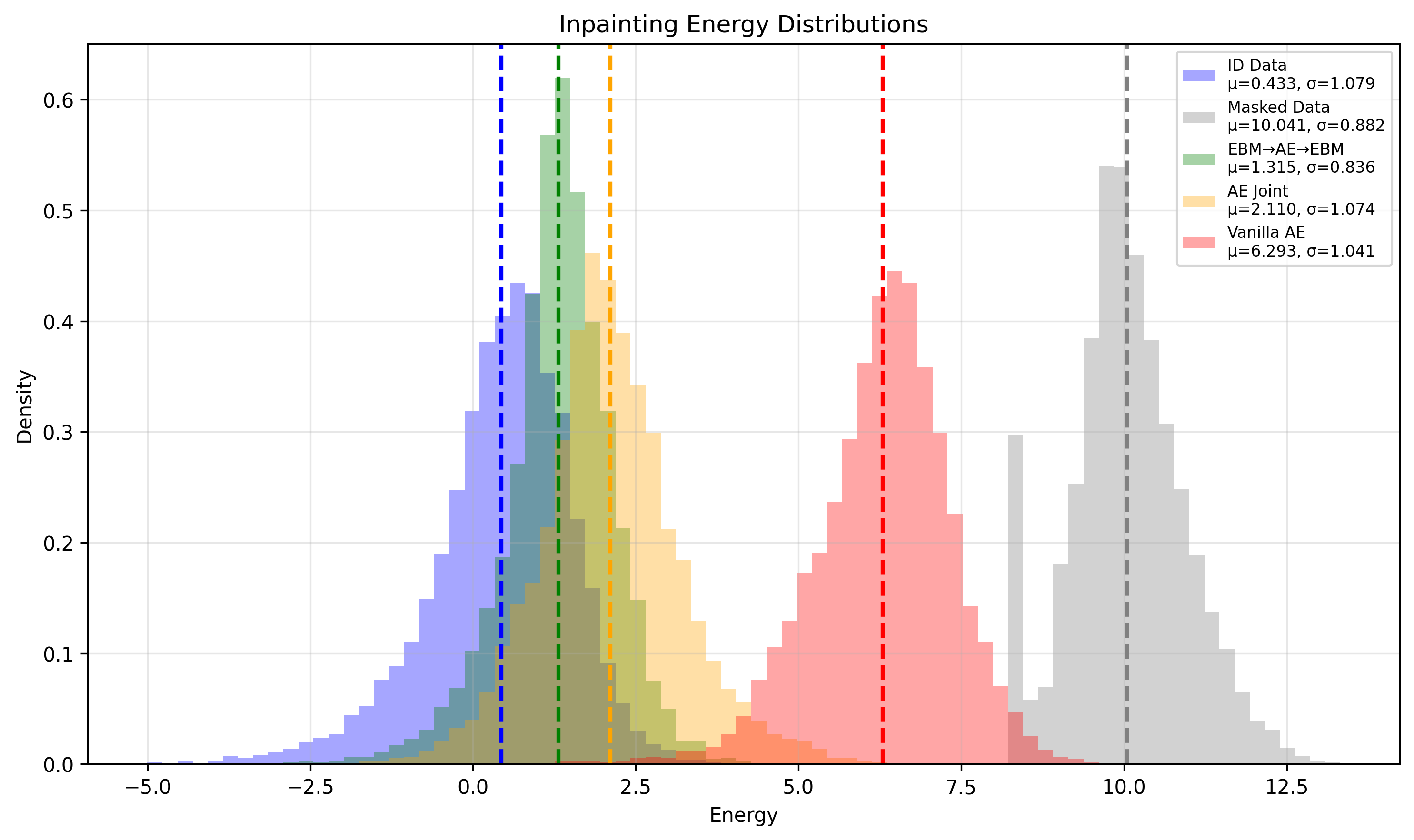} 
\caption{} 
\label{fig:center45_energy} 
\end{subfigure} 

\caption{Image inpainting results obtained for two levels of reconstruction difficulty. The first row corresponds to a center mask covering 40\% of the image, while the second row corresponds to a more challenging scenario where 45\% of the image is removed. For each configuration, the left panel presents representative reconstructions obtained with the different methods, whereas the right panel shows the corresponding energy distributions. For moderate masking ratios (40\%), all methods are able to reconstruct plausible digits, although the proposed EBM$\rightarrow$AE$\rightarrow$EBM framework produces reconstructions that remain closer to the low-energy regions of the learned manifold. As the masking ratio increases to 45\%, the reconstruction problem becomes significantly more difficult and the benefit of the proposed energy-guided refinement becomes more apparent. The EBM$\rightarrow$AE$\rightarrow$EBM framework consistently generates lower-energy and more coherent reconstructions than both the jointly trained AE and the Vanilla AE baseline. } 
\label{fig:inpainting_center_masks} 
\end{figure*}

The EBM$\rightarrow$AE$\rightarrow$EBM framework consistently produces reconstructed images with energies typically ranging between 0.7 and 2.5, regardless of the initial masking configuration. 
In comparison, the Vanilla AE frequently converges toward significantly higher-energy solutions, particularly in the most difficult reconstruction scenarios. The relatively small energy difference between EBM$\rightarrow$AE$\rightarrow$EBM and AE Joint further confirms the conclusions of the previous ablation study. 
Most of the projection toward the learned manifold is already performed by the jointly trained autoencoder. 
The EBM then operates as a complementary refinement mechanism that further minimizes the energy of the reconstructed samples and improves consistency with the learned distribution. 
Figure~\ref{fig:inpainting_center_masks} provides representative reconstruction examples together with the corresponding energy histograms. Two levels of reconstruction difficulty are illustrated. 
The first row corresponds to a moderate masking configuration, whereas the second row presents a substantially more challenging reconstruction task. 
The shift of the energy distributions toward lower values after reconstruction is visually apparent in both scenarios and becomes particularly pronounced for the EBM$\rightarrow$AE$\rightarrow$EBM framework.
For moderate masking ratios, all methods are generally able to reconstruct visually plausible digits. Although some differences can be observed, the reconstructions produced by AE Joint and EBM$\rightarrow$AE$\rightarrow$EBM remain qualitatively similar. 
This observation is consistent with the quantitative results reported in Table~\ref{tab:inpainting_results}, where relatively small performance differences are observed for easier reconstruction tasks. 
The situation changes considerably as the reconstruction problem becomes more difficult. 
In the presence of larger missing regions, the Vanilla AE frequently generates ambiguous or distorted reconstructions, while the jointly trained models preserve a more coherent digit structure. 

\subsection{OOD Detection Performance} 

Beyond image generation and reconstruction, the proposed framework also learns an energy function that can be used for out-of-distribution (OOD) detection. 
Since energy-based models assign low energies to samples that are consistent with the training distribution and higher energies to atypical observations, the learned energy landscape naturally provides a mechanism for discriminating between in-distribution (ID) and OOD data. 
To evaluate this capability, standard OOD detection metrics were monitored throughout training, including the Area Under the Receiver Operating Characteristic Curve (AUROC), the Area Under the Precision-Recall Curve (AUPR), and the False Positive Rate at 95\% True Positive Rate (FPR95). 
Table~\ref{tab:ood_best} reports the best OOD detection results obtained during the experimental campaign. 
The proposed framework consistently achieves strong OOD discrimination performance, with AUROC values ranging from 96.80\% to 98.35\%. Similarly, AUPR values remain above 97\% in all reported experiments, while FPR95 decreases to as little as 1.21\% in the best configuration. These results demonstrate that the learned energy function successfully captures the structure of the MNIST data distribution and provides a reliable indicator of distributional mismatch. Samples lying outside the training manifold are consistently assigned higher energy levels than in-distribution observations, enabling accurate OOD detection. To better understand the role of energy in this process, we additionally analyzed the relationship between replay-buffer energy and OOD performance. 

\begin{table}[ht] 
\centering 
\caption{Best OOD detection scores obtained during training.} 
\label{tab:ood_best} 
\begin{tabular}{lccc} 
\hline 
Experiment & AUROC (\%) & AUPR (\%) & FPR95 (\%) \\
\hline Best Case 1 & 96.80 & 97.81 & 15.75 \\ 
Best Case 2 & 97.40 & 98.28 & 9.76 \\ Best Case 3 & 98.35 & 98.93 & 1.21 \\ 
\hline 
\end{tabular} 
\end{table} 

Table~\ref{tab:energy_ood} reports representative experiments together with their best replay-buffer energy and corresponding AUROC values. 
A clear trend emerges from these results. Experiments associated with strongly negative replay-buffer energies generally exhibit excellent OOD detection performance, with AUROC values consistently exceeding 96\%. 
For example, replay-buffer energies between approximately $-3.2$ and $-5.0$ are systematically associated with strong discrimination capabilities. 
These observations suggest that the energy function progressively organizes the latent space into regions corresponding to realistic and unrealistic samples. Interestingly, the checkpoints achieving the strongest OOD performance do not systematically coincide with the checkpoints producing the best generative performance. 

\begin{table}[ht]
\centering
\caption{Relationship between buffer energy and OOD detection performance.}
\label{tab:energy_ood}
\begin{tabular}{lcc}
\hline
Experiment & Best Buffer Energy & Best AUROC (\%) \\
\hline
test1  & -3.7206 & 98.35 \\
test2  & -3.4898 & 97.91 \\
test3  & -3.3692 & 96.47 \\
test6  & -3.2351 & 97.98 \\
test7  & -3.4062 & 97.48 \\
test9  & -3.5439 & 98.27 \\
test10 & -5.0207 & 96.80 \\
test11 & -3.9724 & 97.35 \\
test12 & -3.8018 & 97.40 \\
\hline
\end{tabular}
\end{table}

\section{Conclusion and Future Perspectives} 

This work introduced a cooperative generative framework that combines an Energy-Based Model (EBM) and a denoising autoencoder within a unified learning architecture. 
The proposed approach is based on an iterative EBM$\rightarrow$AE$\rightarrow$EBM sampling procedure that alternates energy minimization and manifold projection. 
Unlike conventional EBM training strategies that rely exclusively on Langevin dynamics, the proposed framework exploits the complementary strengths of both components. 
The autoencoder learns a projection toward the data manifold, while the EBM refines projected samples according to the learned energy landscape. 

The experimental results consistently support the central hypothesis underlying this work. 
Across all experiments, the jointly trained autoencoder provides the majority of the improvements in generation and reconstruction quality, while the EBM acts primarily as an energy-guided refinement mechanism that further improves consistency with the learned distribution. 
From a geometric perspective, the autoencoder can be interpreted as a nonlinear manifold projection operator that removes sampling artifacts and restores semantic consistency, whereas the EBM performs local energy minimization that drives samples toward low-energy and high-density regions of the learned landscape. 

Extensive experiments conducted on the MNIST dataset demonstrated that joint EBM-AE training substantially improves generation quality compared with a conventional autoencoder. 
The complete EBM$\rightarrow$AE$\rightarrow$EBM framework consistently achieved the best overall balance between generation quality, diversity, and sample validity. 
The ablation study further revealed that the largest performance gain originates from the jointly trained autoencoder itself, while the final energy-refinement stage provides an additional and systematic improvement. 
Beyond unconditional generation, the proposed framework also demonstrated strong image reconstruction and inpainting capabilities. The experimental results showed that corrupted observations initially occupy high-energy regions and are progressively projected toward the learned low-energy manifold. In particular, the benefit of energy-based refinement becomes increasingly apparent as the reconstruction problem becomes more challenging, highlighting the complementary roles of manifold projection and energy minimization.  

Although the present study focuses on image generation and reconstruction, the proposed cooperative learning framework opens several promising research directions. 
In particular, the interpretation of the autoencoder as a manifold projection operator and the EBM as an energy-based refinement mechanism is not restricted to image domains. 
A natural extension of this work concerns time-series forecasting, where predicting future observations can similarly be viewed as searching for low-energy trajectories that remain consistent with the learned temporal manifold. 
Future work will therefore investigate the application of the proposed framework to univariate and multivariate forecasting problems. 
In such a setting, the autoencoder could learn a low-dimensional representation of valid temporal dynamics, while the EBM would evaluate the consistency of predicted trajectories with the historical behavior of the system. 
The cooperative EBM$\rightarrow$AE$\rightarrow$EBM procedure could then be used to iteratively refine future predictions in a manner analogous to the image-generation process studied in this work.

\bibliographystyle{unsrtnat}
\bibliography{references}

\end{document}